\documentclass[sigconf]{acmart}
\usepackage{multirow}
\usepackage{appendix}

\AtBeginDocument{%
  \providecommand\BibTeX{{%
    \normalfont B\kern-0.5em{\scshape i\kern-0.25em b}\kern-0.8em\TeX}}}

\begin{document}

\title{Towards Federated Domain Unlearning: Verification Methodologies and Challenges}

\author{Kahou Tam}
\authornote{Both authors contributed equally to this research.}
\affiliation{%
  \institution{University of Macau}
  \city{Macau SAR}
  \country{China}}
\email{wo133565@gmail.com}

\author{Kewei Xu}
\authornotemark[1]
\affiliation{%
  \institution{University Of Electronic Science And Technology}
  \city{Chengdu}
   \state{Sichuan}
  \country{China}}
\email{2021090909008@std.uestc.edu.cn}

\author{Li Li}
\affiliation{%
  \institution{University of Macau}
  \city{Macau SAR}
  \country{China}}
\email{llili@um.edu.mo}

\author{Huazhu Fu}
\authornote{Corresponding author.}
\affiliation{%
  \institution{Institute of High Performance Computing, Agency for Science, Technology and
Research}
    \city{Singapore}
  \country{Singapore}}
\email{hzfu@ieee.org}


\begin{abstract}

Federated Learning (FL) has evolved as a powerful tool for collaborative model training across multiple entities, ensuring data privacy in sensitive sectors such as healthcare and finance. However, the introduction of the Right to Be Forgotten (RTBF) poses new challenges, necessitating federated unlearning to delete data without full model retraining. Traditional FL unlearning methods, not originally designed with domain specificity in mind, inadequately address the complexities of multi-domain scenarios, often affecting the accuracy of models in non-targeted domains or leading to uniform forgetting across all domains. Our work presents the first comprehensive empirical study on Federated Domain Unlearning, analyzing the characteristics and challenges of current techniques in multi-domain contexts. We uncover that these methods falter, particularly because they neglect the nuanced influences of domain-specific data, which can lead to significant performance degradation and inaccurate model behavior. Our findings reveal that unlearning disproportionately affects the model's deeper layers, erasing critical representational subspaces acquired during earlier training phases. In response, we propose novel evaluation methodologies tailored for Federated Domain Unlearning, aiming to accurately assess and verify domain-specific data erasure without compromising the model's overall integrity and performance. This investigation not only highlights the urgent need for domain-centric unlearning strategies in FL but also sets a new precedent for evaluating and implementing these techniques effectively.
\end{abstract}    
\maketitle

\section{Introduction}

Federated learning (FL) has emerged as a paradigm shift in machine learning, enabling collaborative model training across multiple decentralized entities while preserving data privacy \cite{Konecny2016,Kairouz2019,li2020federated}. This approach is particularly relevant in sectors like healthcare, finance, and telecommunications, where data privacy and security are paramount \cite{Li2020_SPM,tam2023fedcoop,tam2023federated}. By allowing participants to train a shared model without exposing their raw data, federated learning offers a promising solution to the challenges of data silos and privacy concerns.

\begin{figure}[!t]
\vspace{-10pt}
\centering
\includegraphics[width=0.9\linewidth]{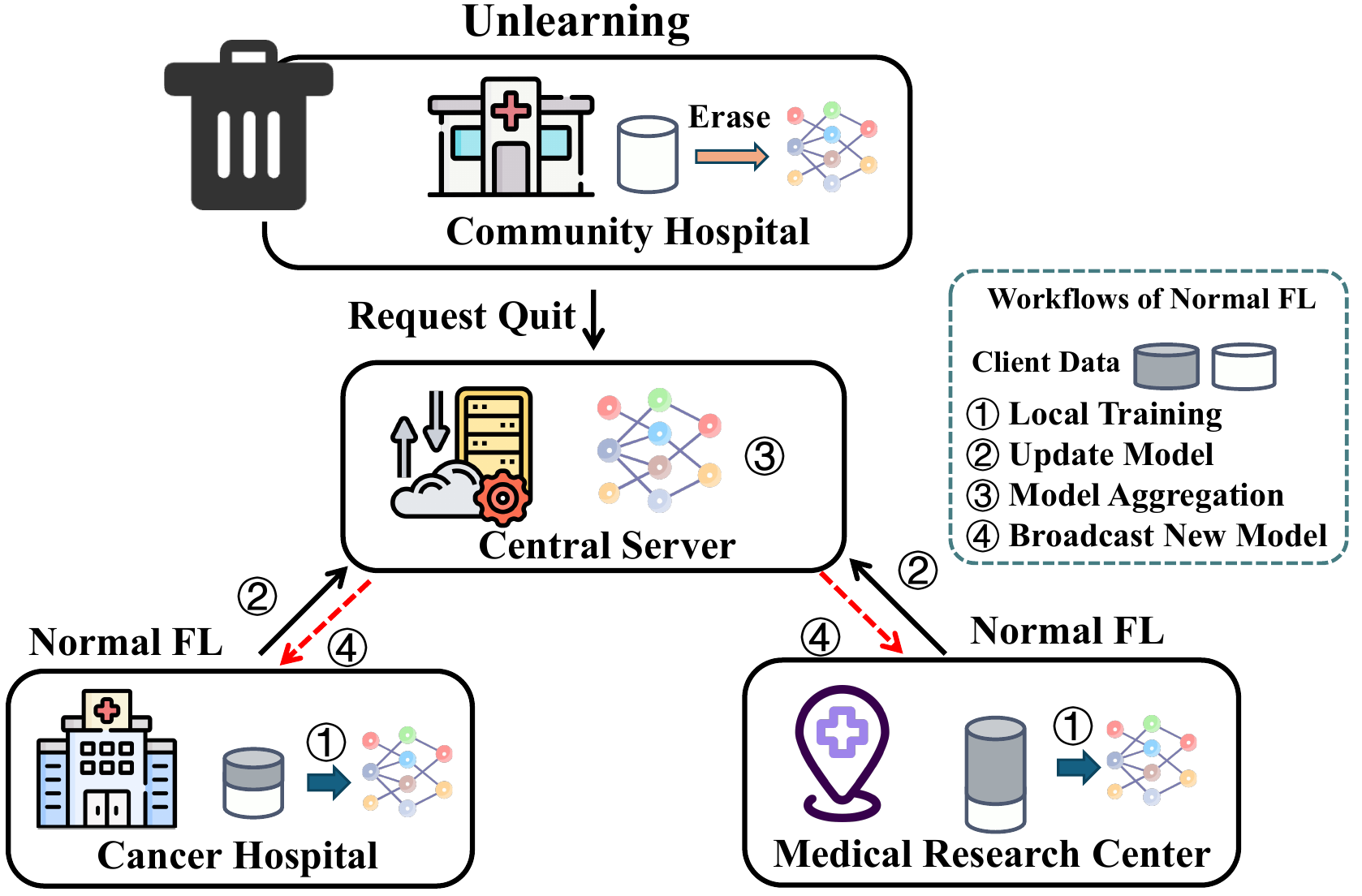}
\vspace{-10pt}
\caption{The workflow of federated unlearning. }
\label{unlearning_framework}
\vspace{-20pt}
\end{figure}

The evolving landscape of data privacy introduces challenges, notably in the context of the Right to Be Forgotten (RTBF) \cite{kalis2014google} within Federated Learning. RTBF mandates the deletion of user data, aiming to preserve model accuracy while complying with privacy regulations such as the General Data Protection Regulation (GDPR) \cite{GDPR2016} and the California Consumer Privacy Act (CCPA \cite{CCPA2020}). To achieve this, an advanced federated scheme, termed "federated unlearning" \cite{liu2023survey,jeong2024sok,liu2021federaser}, has been developed. Figure \ref{unlearning_framework} shows the detailed workflow of federated unlearning. This approach enables the selective erasure of data influences from models without requiring complete retraining. Federated unlearning thus allows organizations to adhere to privacy laws, safeguard user information, and maintain the accuracy and efficiency of FL systems across various sectors. This mechanism is essential for preserving the trustworthiness and integrity of FL systems.

Traditional methods of unlearning, which involve retraining models from scratch, are often impractical due to significant computational demands. Consequently, research in federated unlearning primarily aims to devise methods that efficiently remove data influences and restore model performance within FL's resource-constrained environments \cite{halimi2022federated,wang2022federated,wu2022federated,gao2024verifi}. This process comprises two phases: information removal and performance recovery. Information Removal aims to expunge the influences of targeted data from the trained model, ensuring the model behaves as if it had never encountered the specified data. Techniques include selecting historical information \cite{liu2021federaser}, approximating loss functions \cite{halimi2022federated}, and manipulating gradients\cite{che2023fast}. Following the removal of data influences, it is crucial to restore the global model's performance. Degradation is a common consequence of removing data influences, necessitating recovery strategies. Federated unlearning methods typically employ additional training rounds, knowledge distillation, and fine-tuning with gradient manipulation for performance recovery.

Despite their potential, applying federated unlearning methods in practical scenarios presents significant challenges, primarily due to data heterogeneity \cite{huang2023rethinking,li2020federated}. Traditional federated unlearning approaches predominantly assume that the data distribution among clients is independently and identically distributed (IID) \cite{che2023fast,halimi2022federated,liu2021federaser,gao2024verifi,wu2022federated}. However, this assumption does not hold in many real-world applications, where data heterogeneity is more accurately represented by non-IID distributions. A limited number of studies have explored federated unlearning under non-IID conditions, typically simulating data heterogeneity through imbalanced sampling strategies, such as the Dirichlet distribution \cite{li2020federated}, to create varied label distributions across clients \cite{jeong2024sok}. Nonetheless, the scenario of multi-domain FL, where client data stem from varied domains, remains underexplored despite its real-world relevance \cite{huang2023rethinking,li2021fedbn}. In practical applications, private data can emanate from diverse domains, leading to significantly different feature distributions. For instance, autonomous vehicles collaborating on model training might gather data under varying weather conditions or times of the day, resulting in domain disparities in the collected images. Similarly, healthcare institutions participating in medical imaging analysis may encounter substantial domain gaps due to differences in imaging equipment and protocols. Addressing the challenge of domain heterogeneity in FL, recent studies have explored innovative approaches. Prototype Learning has emerged as a method to impart valuable domain knowledge and establish a fair convergent target for global model training \cite{huang2023rethinking,huang2022few}. These methods enable the global model to assimilate general knowledge across diverse domains by distilling specific domain knowledge, thereby enhancing model robustness and applicability in heterogeneous data environments.

The exploration of federated unlearning within a multi-domain FL framework shows promising potential but also significant challenges due to diverse domain data among clients. Traditional unlearning methods \cite{gao2024verifi,halimi2022federated,liu2021federaser,wu2022federated,che2023fast,jeong2024sok}, designed for homogeneous or mildly heterogeneous settings, struggle with the complexities of multi-domain data distribution, raising concerns about their effectiveness. The primary challenge is the nuanced task of precisely identifying and removing domain-specific influences without impairing the global model's performance or requiring excessive computational resources. Additionally, maintaining the model's integrity and performance across all domains post-unlearning requires sophisticated strategies. The evaluation of unlearned models in this context calls for innovative and comprehensive approaches, incorporating evaluation metrics that assess not only accuracy and loss but also domain-specific performance, the preservation of utility across non-targeted domains, and the thorough removal of information from the target domain. These issues underscore the need to deepen our understanding of federated unlearning in multi-domain environments and to develop tailored novel verification methodologies.

In this study, we delve into the nuanced relationship between the processes of unlearning and the specificity of domains within multi-domain FL frameworks. Our work introduces an innovative evaluation methodology for assessing the effectiveness of unlearned models in such complex FL environments. Initially, we examine the performance of current federated unlearning approaches within multi-domain contexts. Our thorough empirical analysis indicates that existing techniques, predominantly developed for single-domain scenarios, inadequately address the unique challenges of multi-domain FL. These conventional methods often either degrade the model accuracy in ancillary domains or lead to indiscriminate data forgetting across all domains when targeting specific data for removal. Through detailed quantitative analysis, we uncover that the model's deeper layers predominantly facilitate forgetting, effectively wiping clean earlier learned representational spaces. 
Motivated by these insights, we propose novel validation techniques specifically tailored for federated domain unlearning. Our approach involves selecting a subset of representative samples of the unlearned domain and employing a generative model to introduce adversarial noise as the marker into these samples. This marker is then combined with the original representative samples to create 'confusing' samples aimed at deceiving the local model in the unlearned domain. To enhance the practicality and effectiveness of this verification method, we impose constraints on the unlearned domain's model update, ensuring that it remains within the vicinity of the global model. This constraint helps bind the samples with markers close to the domain-specific knowledge, thereby providing a more accurate assessment of the unlearning process's efficacy.
Specifically, we make the following key contributions: 

$\bullet$ We introduce the first empirical study on Federated Domain Unlearning, in which we meticulously analyze the complexities and challenges faced by current unlearning techniques across varied domain contexts to guide the development of more robust approaches.

$\bullet$ We identify and detail the critical shortcomings of prevailing unlearning methods, particularly highlighting their neglect of domain-specific data nuances, which significantly impairs model performance and behavior. Our findings prompt the necessity for enhanced methodological precision and sensitivity.

$\bullet$ To address these challenges, we develop and validate innovative verification methodologies tailored for Federated Domain Unlearning. These methods are designed to precisely assess and confirm the successful erasure of domain-specific data without undermining the overall integrity and efficacy of the learning model.

\section{Related Work}

\subsection{Federated Learning with Domain Heterogeneity}

In addressing domain heterogeneity in Federated Learning, the literature reveals two primary methodological camps: Prototype Learning \cite{huang2023rethinking,huang2022few} and Domain Adaptation \cite{huang2022learn,zhang2023federated}. Prototype Learning approaches focus on abstracting domain-specific features into more generalizable prototypes, facilitating a more effective model learning across varied domains \cite{huang2023rethinking,huang2022few}. These methods often leverage the concept of prototype representation to reduce the impact of domain variability, thus enhancing model robustness against domain heterogeneity \cite{huang2023rethinking}. However, while these strategies excel in abstracting domain features, they primarily concentrate on optimizing model performance within multi-domain contexts, without delving into the complexities of unlearning domain-specific biases. On the other hand, Domain Adaptation techniques aim to bridge the gap between different domains by aligning the feature spaces or distributions, ensuring that the model trained on one domain can perform well on others \cite{huang2022learn,zhang2023federated}. These approaches often employ strategies like adversarial training, domain feature alignment, and transfer learning to mitigate the effects of domain heterogeneity. Despite their effectiveness in handling domain shifts, these techniques similarly focus on the adaptation aspect, primarily concerned with how to learn across multiple domains effectively. They lack a focus on unlearning, which involves actively removing domain-specific knowledge that might hinder generalization to new, unseen domains.

While both Prototype Learning and Domain Adaptation offer promising avenues for tackling domain heterogeneity, they primarily concentrate on the aspect of learning from multi-domain data. This focus overlooks the potential benefits of unlearning, which involves intentionally forgetting specific data or features that could lead to overfitting or bias towards certain domains. Incorporating unlearning mechanisms could significantly enhance the flexibility and robustness of Federated Learning models by enabling them to discard non-essential or detrimental information, thus focusing learning efforts on truly relevant and generalizable knowledge.

\subsection{Federated Unlearning}

In federated learning, data is distributed across multiple devices or nodes and is not centrally stored, presenting unique challenges for unlearning. As a result, federated unlearning methods predominantly focus on approximate unlearning. These methods can be further categorized into three types:

$\bullet$ \textbf{Retrain Unlearning} adapts the exact unlearning approach from centralized learning but forgoes the partition-aggregation framework, which is not effective in improving time efficiency in the federated context. Instead, rapid retrain unlearning \cite{liu2021federaser,liu2022right} methods seek to accelerate the retraining process by approximating the gradients using previously stored historical updates. However, this approach faces challenges due to the potential inaccuracies in gradient approximation and the additional burden of data storage for maintaining historical updates.

$\bullet$ \textbf{Reverse Unlearning} removal of the learned data through reverse gradient operations, e.g., loss maximization \cite{halimi2022federated} and Gradient Manipulation \cite{wu2022federated,che2023fast}. Loss functions only on the remaining data are approximated by calculating the inverse Hessian matrix as if the unlearned model has not been trained on the target data. Gradient Manipulation injects the noise to smooth all local models’ gradients, treating them as perturbations during server
aggregation. 

$\bullet$ \textbf{Other.} There are other federated unlearning methods using knowledge distillation \cite{wu2022federated2}, scaled gradients \cite{gao2024verifi}, and channel pruning (CNN-specific) \cite{wang2022federated}. The theoretical underpinnings of these methods are notably weaker than those of the above two approaches, placing them at a greater risk of encountering privacy concerns.

In federated unlearning, the granularity of the unlearning target can be classified into four categories \cite{jeong2024sok,liu2023survey}: sample-wise, client-wise, class-wise, and feature-level. \textbf{Sample-wise} unlearning involves removing specific data samples from individual clients' datasets. This is commonly requested when certain data points are identified as erroneous or sensitive and need to be excluded from the model's training process.
\textbf{Client-wise} unlearning focuses on eliminating the entire dataset or influence of a specific client from the federated model. This could be necessary if a client withdraws from the federation or if their data is found to be compromised or biased.
\textbf{Class-wise} unlearning aims to remove the influence of a particular class or category across all clients in the federation. This might be required if a specific category becomes irrelevant or is deemed inappropriate for the model's use case.
\textbf{Feature-level} unlearning entails modifying or removing certain features or attributes across the datasets of all clients. This is often essential in scenarios where specific features are identified as sensitive or potentially leading to biased model outcomes. However, Domain-wise unlearning does not directly fit into the previously mentioned categories of sample-wise, client-wise, class-wise, or feature-level unlearning in federated settings. While feature-level unlearning focuses on specific attributes or characteristics within the data, domain-wise unlearning deals with the broader concept of data domains or distributions.  Domain-wise unlearning is concerned with removing the influence of an entire domain or distribution from the federated model. In federated learning, a domain could represent a distinct group of clients sharing similar characteristics or data distributions. For example, in a healthcare federated learning scenario, one domain might represent hospitals in a particular geographic region, while another domain could represent clinics specializing in a specific medical field.

\section{Federated Domain Unlearning}

\noindent\textbf{Notations.} Following the standard federated learning setup \cite{huang2023rethinking,li2020federated,mcmahan2017communication,liu2021federaser}, there are $M$ clients (index $i$) with respective private data $D_i$ containing scale of $N_i$ for the $i$ client private dataset. For each client's private data $D$, each data sample is represented as $(x,y)$, where $x$ is the input attribute and $y$ is the label. In federated learning with heterogeneous domain data, the conditional feature distribution $P(x|y)$ varies across participants even if $P(y)$ is consistent. As for the same label space, distinct feature distribution exists among different participants \cite{li2021fedbn,huang2022few,huang2023rethinking}, resulting in domain shift: 
\begin{equation}
    P_i(x|y)\neq P_j(x|y)  \quad \text{where} \quad P_i(y)==P_j(y)
\end{equation}
Besides, for each client's local training, we regard the model as consisting of two parts: feature extractor $f$ and classifier $g$. The feature extractor encodes sample $x$ into compact $d$ dimensional feature vector $z=f(x) \in \mathbb{R}^d$ in the feature space $Z$. Then the classifier $g$ maps the feature $z$ into logits output $l= g(z)$. The goal of FL is to collaboratively learn a machine learning model, $w$ to minimize the weighted empirical loss among clients:
\begin{equation}
    w= \text{min} \sum^M_{i=1} \frac{D_i}{\sum_{i=1}^{M} D_i } F_i(w,D_i)
\end{equation}
where $F_i(w,D_i)$ represents the empirical loss.

\noindent\textbf{Problem Formulation of Federated Domain Unlearning.}
Considering the notations and setup described earlier, the objective of federated domain unlearning is to efficiently remove the influence of a specific client's domain data from the federated model while preserving the model's performance on the remaining clients \cite{huang2023rethinking}. Let's denote the client whose domain is to be unlearned as $k$. The goal of federated domain unlearning can be formulated as follows:

\noindent$\bullet$ \textbf{Domain Removal}: Remove the domain-specific information of client $k$ from the federated model. This involves updating the model parameters in a way that the model no longer retains any information specific to the domain of client $k$. Formally, we aim to find a new model $w'$ such that:
\begin{equation}
d(f'(x), f(x)) \geq \delta, \quad \forall x \in D_k
\end{equation}
where $d(\cdot, \cdot)$ is a distance metric, and $\delta$ is a threshold indicating a significant difference in the model's response to the data from the domain to be unlearned.

\noindent$\bullet$ \textbf{Model Preservation}: Ensure that the updated model $w'$ maintains its performance on the remaining clients' domains. This can be expressed as:
\begin{equation}
\left| F_i(w', D_i) - F_i(w, D_i) \right| \leq \alpha, \quad \forall i \neq k
\end{equation}
where $\alpha$ is a small positive value indicating the acceptable deviation in performance.

\subsection{Comparison with Prior Federated Unlearning}
The concept of unlearning in FL has been previously explored in various contexts, such as data unlearning \cite{che2023fast,halimi2022federated} and class unlearning \cite{wang2022federated}. However, federated domain unlearning introduces a nuanced perspective by focusing on the removal of domain-specific information while preserving the model's generalization ability across the remaining domains.

\textbf{Objective Function Comparison:} The objective function in federated domain unlearning, as formulated earlier, involves minimizing the distance between the updated model \(f'\) and a model \(f_{-k}\) trained without the data from the target domain. This is distinct from typical federated unlearning objectives, which may focus solely on minimizing the impact of removed data points \cite{che2023fast,liu2023survey}. The additional constraint in federated domain unlearning, which ensures that the performance on the remaining clients' data is similar to the original model, further differentiates it from conventional unlearning approaches.

\textbf{Generalization Ability:} A key aspect of federated domain unlearning is its emphasis on preserving the model's generalization ability across the remaining domains \cite{huang2023rethinking}. This is crucial in federated settings where data heterogeneity is common. By ensuring that the unlearned model maintains its performance on other clients' data, federated domain unlearning addresses the challenge of domain shift \cite{halimi2022federated,li2021fedbn}, which is often overlooked in traditional unlearning methods.

\section{Empirical Study}

\begin{table*}[!t]
\vspace{-7pt}
\caption{
Evaluation of Federated Domain Unlearning Across Various Methods on DomainNet dataset.}
\vspace{-10pt}
\label{result1}
\renewcommand{\arraystretch}{1.2}
\resizebox{\linewidth}{!}{%

\begin{tabular}{cccccccccccccccccc}
\hline
 &
   &
  Train &
  \multicolumn{6}{c}{Test  ACC  For All  Domain} &
  &
   &
  Train &
  \multicolumn{6}{c}{Test  ACC  For All  Domain} \\ \hline
\multirow{2}{*}{\begin{tabular}[c]{@{}c@{}}Unlearn\\  Domain\end{tabular}} &
  Method &
  / &
  Clipart &
  Infograph &
  Painting &
  Quickdraw &
  Real &
  Sketch &
  \multirow{2}{*}{\begin{tabular}[c]{@{}c@{}}Unlearn \\ Domain\end{tabular}} &
  Method &
  / &
  Clipart &
  Infograph &
  Painting &
  Quickdraw &
  Real &
  Sketch \\  \hline
 &
  Full learn &
  98.15±0.56 &
  86.69±0.51 &
  49.01±0.78 &
  78.35±0.35 &
  76.60±0.69 &
  79.62±0.34 &
  84.48±0.55 &
   &
  Full learn &
  98.15±0.56 &
  86.69±0.51 &
  49.01±0.78 &
  78.35±0.35 &
  76.60±0.69 &
  79.62±0.34 &
  84.48±0.55 \\  \hline
\multirow{5}{*}{Clipart} &
  Retrain &
  68.19±0.29 &
  72.43±0.64 &
  45.97±1.02 &
  77.54±0.56 &
  72.45±0.37 &
  78.88±0.43 &
  81.05±0.76 &
  \multirow{5}{*}{Quickdraw} &
  Retrain &
  50.45±1.55 &
  84.98±1.54 &
  47.03±1.44 &
  81.10±0.55 &
  51.72±0.59 &
  82.83±0.14 &
  84.12±0.56 \\
 &
  BL1 RR &
  51.88±1.85 &
  56.46±1.64 &
  32.27±1.55 &
  54.28±1.95 &
  63.16±0.86 &
  61.54±1.57 &
  50.18±1.56 &
   &
  BL1 RR &
  40.51±2.50 &
  69.39±0.94 &
  34.70±1.14 &
  62.84±1.34 &
  41.21±0.74 &
  71.82±0.54 &
  59.93±0.86 \\
 &
  BL2 FE &
  67.38±1.64 &
  71.48±1.22 &
  40.64±1.37 &
  72.54±0.59 &
  66.51±0.44 &
  74.94±0.67 &
  77.98±0.54 &
   &
  BL2 FE &
  48.60±1.25 &
  81.18±0.34 &
  42.62±1.26 &
  76.74±0.98 &
  48.80±2.98 &
  76.17±0.96 &
  77.62±0.28 \\
 &
  BL3 IL &
  81.93±0.29 &
  78.14±1.24 &
  45.51±0.18 &
  80.45±0.55 &
  72.82±0.72 &
  79.38±1.30 &
  86.12±0.15 &
   &
  BL3 IL &
  54.73±0.34 &
  83.27±0.57 &
  47.95±1.25 &
  77.38±054 &
  55.17±0.45 &
  78.88±0.47 &
  80.87±0.35 \\
 &
  BL4 CP &
  76.18±0.75 &
  77.76±0.37 &
  45.05±1.52 &
  79.16±0.42 &
  75.05±0.97 &
  80.12±0.37 &
  84.48±0.67 &
   &
  BL4 CP &
  54.78±0.72 &
  86.69±0.15 &
  49.47±0.54 &
  83.36±0.63 &
  56.25±0.76 &
  82.91±0.85 &
  86.82±0.45 \\ \hline
\multirow{5}{*}{Infograph} &
  Retrain &
  36.04±0.54 &
  86.50±0.42 &
  37.14±0.27 &
  79.32±0.37 &
  76.91±0.86 &
  80.44±0.45 &
  86.11±0.37 &
  \multirow{5}{*}{Real} &
  Retrain &
  72.94±0.34 &
  85.55±0.57 &
  45.81±0.64 &
  75.28±0.97 &
  76.44±0.67 &
  71.98±0.54 &
  86.13±0.24 \\
 &
  BL1 RR &
  28.01±0.94 &
  68.82±0.43 &
  28.46±0.35 &
  58.00±0.65 &
  63.14±0.34 &
  62.94±0.81 &
  61.73±0.43 &
   &
  BL1 RR &
  55.16±0.56 &
  71.67±0.45 &
  32.42±0.38 &
  53.63±1.54 &
  70.42±1.54 &
  54.89±1.00 &
  56.86±0.47 \\
 &
  BL2 FE &
  33.48±1.35 &
  84.03±0.45 &
  35.01±0.35 &
  76.74±0.67 &
  75.82±0.84 &
  79.05±0.67 &
  82.49±0.21 &
   &
  BL2 FE &
  71.37±0.83 &
  83.84±0..24 &
  42.62±1.42 &
  74.64±0.35 &
  71.31±0.45 &
  70.99±1.15 &
  81.41±1.54 \\
 &
  BL3 IL &
  70.00±0.54 &
  87.83±024 &
  43.53±1.34 &
  80.29±0.35 &
  81.66±1.22 &
  83.41±0.53 &
  85.74±0.44 &
   &
  BL3 IL &
  81.93±0.54 &
  84.60±0.45 &
  43.07±0.44 &
  76.74±0.37 &
  77.20±0.51 &
  75.76±0.54 &
  84.48±0.62 \\
 &
  BL4 CP &
  37.38±1.54 &
  89.16±021 &
  39.42±1.36 &
  81.10±0.54 &
  78.21±1.37 &
  82.99±0.34 &
  88.09±0.52 &
   &
  BL4 CP &
  74.47±1.24 &
  84.98±0.75 &
  47.64±0.34 &
  78.03±0.61 &
  79.22±0.94 &
  72.97±0.87 &
  87.36±0.46 \\  \hline
\multirow{5}{*}{Painting} &
  Retrain &
  65.57±1.41 &
  85.17±0.45 &
  45.21±1.11 &
  69.79±1.24 &
  75.93±0.12 &
  80.77±0.43 &
  81.23±044 &
  \multirow{5}{*}{Sketch} &
  Retrain &
  63.15±1.15 &
  81.56±0.34 &
  43.99±0.84 &
  75.28±0.87 &
  70.55±0.43 &
  78.82±0.87 &
  66.43±0.78 \\
 &
  BL1 RR &
  49.07±1.86 &
  71.10±1.25 &
  33.64±1.67 &
  47.50±2.45 &
  66.42±1.24 &
  63.93±1.54 &
  58.35±2.87 &
   &
  BL1 RR &
  39.37±1.52 &
  65.59±1.54 &
  30.44±1.56 &
  52.83±0.97 &
  63.47±0.76 &
  59.98±1.00 &
  40.61±1.54 \\
 &
  BL2 FE &
  65.00±1.25 &
  84.79±1.97 &
  43.68±1.25 &
  70.11±2.03 &
  72.91±1.00 &
  76.99±0.94 &
  79.96±0.75 &
   &
  BL2 FE &
  58.16±0.45 &
  77.57±1.76 &
  42.62±0.43 &
  72.21±0.50 &
  68.10±0.53 &
  73.54±0.79 &
  64.44±0.87 \\
 &
  BL3 IL &
  85.69±0.45 &
  87.07±0.12 &
  46.42±1.96 &
  74.47±1.15 &
  81.3.5±0.96 &
  81.76±0.45 &
  85.56±0.76 &
   &
  BL3 IL &
  89.92±0.87 &
  86.69±0.64 &
  49.47±0.67 &
  81.12±0.34 &
  78.01±0.78 &
  80.36±0.71 &
  81.41±0.76 \\
 &
  BL4 CP &
  70.42±1.25 &
  87.64±1.24 &
  46.58±1.25 &
  71.73±1.45 &
  75.91±0.25 &
  80.61±0.22 &
  85.38±0.74 &
   &
  BL4 CP &
  73.89±0.97 &
  84.60±0.37 &
  47.95±0.85 &
  78.19±0.73 &
  73.41±1.35 &
  80.03±0.78 &
  78.16±0.45 \\ \hline
\end{tabular}
}
\vspace{-15pt}
\end{table*}

\subsection{Setup}
\textit{1) Datasets:} We conduct experiments using three datasets, including Domain-Digits \cite{hull1994database,lecun1998gradient,netzer2011reading,roy2018effects,ganin2015unsupervised} and Office-Caltech \cite{gong2012geodesic}, and DomainNet \cite{peng2019moment}. Each dataset contains data from different domains with domain heterogeneity, but having the same labels. Each federated client is assigned data from one domain.

\textit{2) Neural Network Architectures:} 
During the federated learning training and federated unlearning process, both the clients and server utilize the same model. For different datasets, we employ distinct networks to perform the classification tasks. For Domain-Digits, we use the model consisting of 3 convolution layers, 2 max-pool layers and 3 fully connected layers as previous works \cite{li2021fedbn}. As for Office-Caltech and DomainNet, we use VGG16.

\textit{3) FL Settings:}
During the federated learning process, for each dataset, we assign an entire domain of data to each client. The local update epoch is set to 10, and the global train rounds are 50 for all datasets. We use the cross-entropy loss function and an SGD optimizer with a learning rate of 0.01 for local updates. Before the unlearning,  we utilize the state-of-the-art cluster based Federated Prototypes Learning (FPL) \cite{huang2023rethinking} to train the global model among clients with diverse domain data. All the hyper-parameters are followed by the original work \cite{huang2023rethinking}.

\textit{4) Federated unlearning Method: }

In our study, we evaluated five advanced federated unlearning methods, which can be categorized into three major types. The first category, retrain learning, includes models such as Retrain, which involves retraining from scratch while excluding the data of the participant to be forgotten; Rapid Retraining \cite{liu2022right}, a swift approach designed to entirely erase data samples from a trained Federated Learning (FL) model; and FedEraser \cite{liu2021federaser}, a method that efficiently removes the impact of a client's data on the global FL model while significantly reducing the time required for constructing the unlearned FL model. The second category, reverse training, is represented by the Increase Loss \cite{halimi2022federated}, which performs unlearning at the client level by reversing the learning process, i.e., training a model to maximize the local empirical loss. The third category, other methods, includes Class-Discriminative Pruning \cite{wang2022federated}, which employs CNN channel pruning to guide the federated machine unlearning process, selecting channels based on TF-IDF scores.

\subsection{Effectiveness of Existing Methods in Federated Domain Unlearning}
We conduct an empirical evaluation of the effectiveness of current unlearning methods in the context of multiple domains. The unlearned domain's accuracy and the remaining domains' test accuracy for the DomainNet dataset are presented in Tables \ref{result1}. More results from other datasets are shown in supplementary materials. The "Retrain" method serves as a benchmark for comparing the efficacy of these existing unlearning approaches. 
In the exploration of federated domain unlearning, the methodologies applied yield diverse impacts on the testing accuracies of various domains. The Retrain method, when excluding certain domains, inevitably results in variability of testing accuracy declines among remaining domains due to differences in the complexity and contributions of each domain during training. Conversely, Rapid Retraining reaches similar outcomes for the forgotten domains but adversely affects the learning of the remaining domains, reducing their testing accuracy by 0.5\%-5\% in the domain-digital dataset and 5\%-30\% in DomainNet. This reduction in performance stems from the method's failure to differentiate adequately between domain-specific and generalized knowledge during retraining.
Similarly, the FedEraser technique, while effectively removing unwanted domain knowledge, inadvertently leads to significant unlearning in the domains meant to be retained, with testing accuracy drops as high as 12\% in the domain-digital dataset and 10.1\% in DomainNet. This issue is likely due to the overly aggressive erasure of critical shared features across domains, undermining the model's overall effectiveness.
Contrastingly, the Increase Loss method does not effectively forget the targeted domains, maintaining or even increasing their accuracy. For example, in DomainNet, the accuracy in the Sketch domain was significantly higher compared to that achieved by Retrain up to 35\%, indicating its failure to adequately perform targeted unlearning while not affecting other domains.
Similarly, Class-Discriminative Pruning also struggles to effectively erase the knowledge related to the domains intended to be forgotten. This results in the accuracy for these domains being up to 10\% higher in DomainNet compared to Retrain, albeit not to the extent observed with the Increase Loss method. This method's limitation lies in its reliance on CNN channel pruning guided by TF-IDF scores, which might not accurately capture the relevance of channels to specific domains, leading to incomplete or ineffective pruning of domain-specific features.

In summary, the current methodologies for federated unlearning present significant challenges in the context of federated domain unlearning. These methods tend to either compromise the learning of original domains while attempting to forget targeted domains or fail to precisely erase the information of the targeted domains. This dichotomy highlights a fundamental limitation in existing approaches, where the balancing act between effectively unlearning specific domain data and preserving the integrity and performance of non-targeted domains remains unresolved. The inability to selectively forget without residual impact underscores the need for more refined techniques that can manage domain-specific unlearning without undermining the overall federated learning system's effectiveness and robustness.

\begin{figure}[!t]
\centering
\includegraphics[width=1\linewidth]{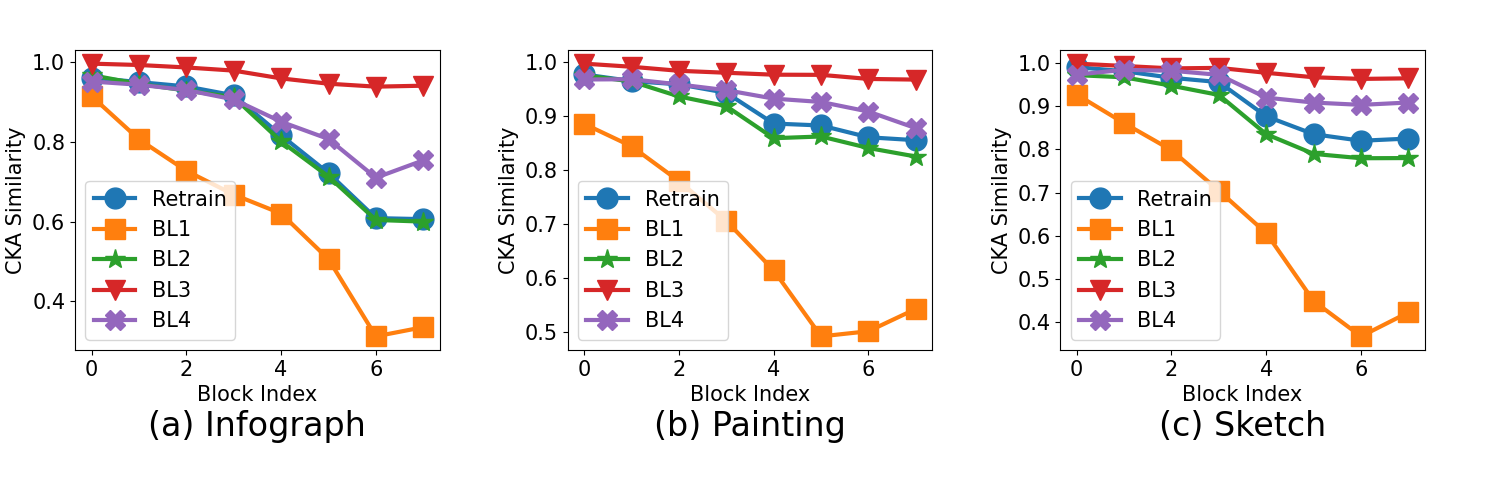}
\vspace{-28pt}
\caption{CKA Analysis of Layer Representations Before and After Unlearning the Target Domain in DomainNet. We select three domains to display: (a) Infograp, (b) Painting, and (c) Sketch. More detail results are shown in supplementary materials. }
\label{fig_cka}
\vspace{-24pt}
\end{figure}

\begin{figure}[!t]
\centering
\includegraphics[width=0.9\linewidth]{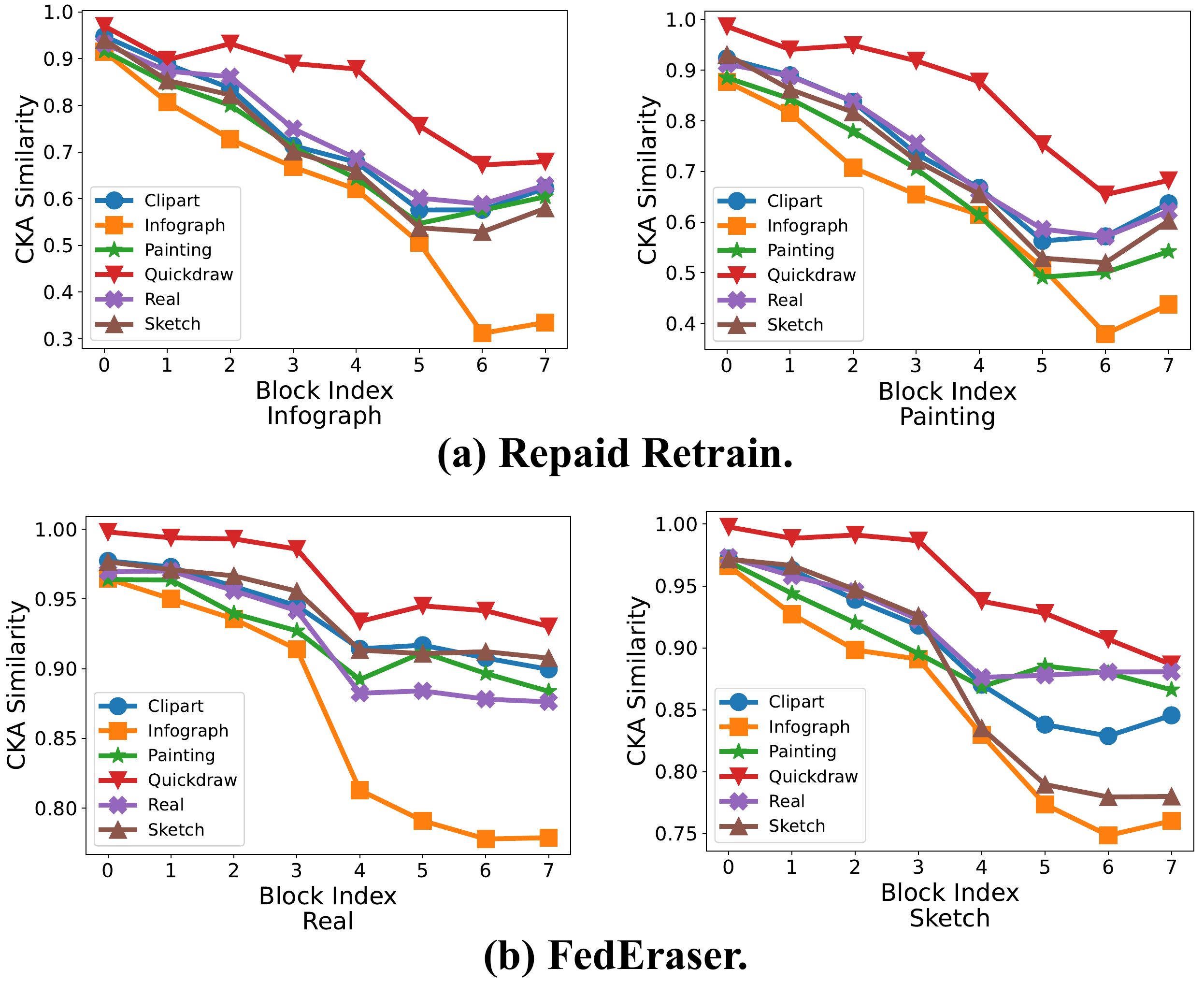}
\vspace{-10pt}
\caption{Comparative CKA Analysis of Layer Representations in Unlearned and Remaining Domains in DomainNet. We report the results of the methods Repaid-Retrain and FedEraser, which unlearn the target domain but also impact the remaining domain's integrity.}
\label{fig_cka_remain_compare}
\vspace{-26pt}
\end{figure}

\subsection{Federated Domain Unlearning and Hidden Layer Representations}
To address the observed shortcomings in federated domain unlearning methods, we undertake an in-depth investigation into the hidden neural network representations during the unlearning process. The focus is on understanding the dynamics of target domain forgetting and the preservation of remaining domain knowledge. This investigation is structured around two pivotal questions:
\textit{(1) Do all parameters (and layers) forget equally with respect to the target domain?}
\textit{(2) Is the knowledge pertaining to the remaining domains preserved after the target domain has been unlearned?}

To explore these questions, we employ the Centered Kernel Alignment (CKA) metric, a tool for assessing the similarity between neural network representations. Originally introduced by Kornblith et al. \cite{kornblith2019similarity}, CKA quantifies the similarity between two neural networks by computing the inner product between their centered kernel matrices. This approach provides a measure of how much common information is retained between the networks, thereby shedding light on the extent of information preservation or loss during unlearning.
In our experimental setup, we utilize linear CKA to analyze the similarity of the output features produced by two models before and after the unlearning process. Given a dataset \(D_{cka}\), we extract feature matrices \(Z_1\) and \(Z_2\) from the two models, respectively. The linear CKA similarity between two representations \(X\) and \(Y\) is calculated using the following equation:
\[
CKA(X,Y) = \frac{||X^TY||^2_F}{||X^TX||^2_F \cdot ||Y^TY||^2_F},
\]
where \(|| \cdot ||_F\) denotes the Frobenius norm. This formula yields a similarity score ranging from 0 (indicating no similarity) to 1 (indicating identical representations), thereby enabling a quantitative assessment of how similar the output features of the same layer are across two models.

Through this analytical approach, we aim to deepen our understanding of the neural mechanisms involved in the unlearning process by comparing the original model with models that have undergone domain-specific unlearning. Figure \ref{fig_cka} displays the results of Centered Kernel Alignment (CKA) across multiple target domains from the DomainNet dataset, comparing various unlearning methods with the comprehensive learning model. More results from other datasets are shown in supplementary materials. Notably, methods like Rapid Retraining and FedEraser, which effectively unlearn target domains, show significant decreases in representation similarity in higher layers, indicating a pivotal role in erasing domain-specific knowledge. In contrast, methods such as Increase Loss and Class-Discriminative Pruning exhibit minimal variation in CKA scores across layers, suggesting a lack of sensitivity and specificity to the targeted domains and a failure to effectively erase domain-specific knowledge. Their representations remain closely aligned with the full learning model, highlighting their limitations in modifying domain-specific information.

Additionally, while methods like Rapid Retraining and FedEraser are effective at erasing target domain knowledge, they inadvertently affect the learning of remaining domains. Figure \ref{fig_cka_remain_compare} shows that while there is a discernible decrease in similarity for the target domain with deeper layers, indicating some unlearning, other domains maintain more stable similarity scores. Notably, methods that successfully lower the CKA scores for unlearned domains often also decrease scores for some remaining domains, illustrating the complexity of domain interdependencies within neural representations and the challenge of achieving precise unlearning without impacting related domains. For instance, while FedEraser unlearns the "Real" domain, it also disrupts the "Infograph" domain, leading to reduced CKA scores and highlighting the need for more refined unlearning methods to avoid such collateral effects.

\subsection{Feature Reuse}
To further investigate how the representations of lower and higher layers evolve during unlearning, we conduct the subspace similarity analysis on the unlearned models with different unlearning methods. Let \( A \in \mathbb{R}^{n \times m} \) represent the centered layer activation matrix with \( n \) examples and \( m \) neurons. We determine the PCA decomposition of \( A \), which involves computing the eigenvectors \( (e_1, e_2, ...) \) and the corresponding eigenvalues \( (\delta_1, \delta_2, ...) \) of the matrix \( A^TA \). Let \( E_k \) denote the matrix composed of the first \( k \) principal components, with \( e_1, ..., e_k \) as its columns, and let \( G_k \) be the analogous matrix derived from another activation matrix \( B \). We then compute the subspace similarity for the top \( k \) components as:
\vspace{-6pt}
\begin{equation}
    \text{SubspaceSim}_k(A, B) = ||G_k^T \cdot E_k||_F^2 
    \vspace{-4pt}
\end{equation}
This metric quantifies the congruence of the subspaces spanned by \( (e_1, ..., e_k) \) and \( (g_1, ..., g_k) \). For instance, if \( A \) and \( B \) are the layer activation matrices corresponding to different tasks, then \( \text{SubspaceSim}_k \) evaluates the similarity in how the network encodes the top \( k \) features for those tasks. 

\begin{figure}[!t]
\centering
\includegraphics[width=1\linewidth]{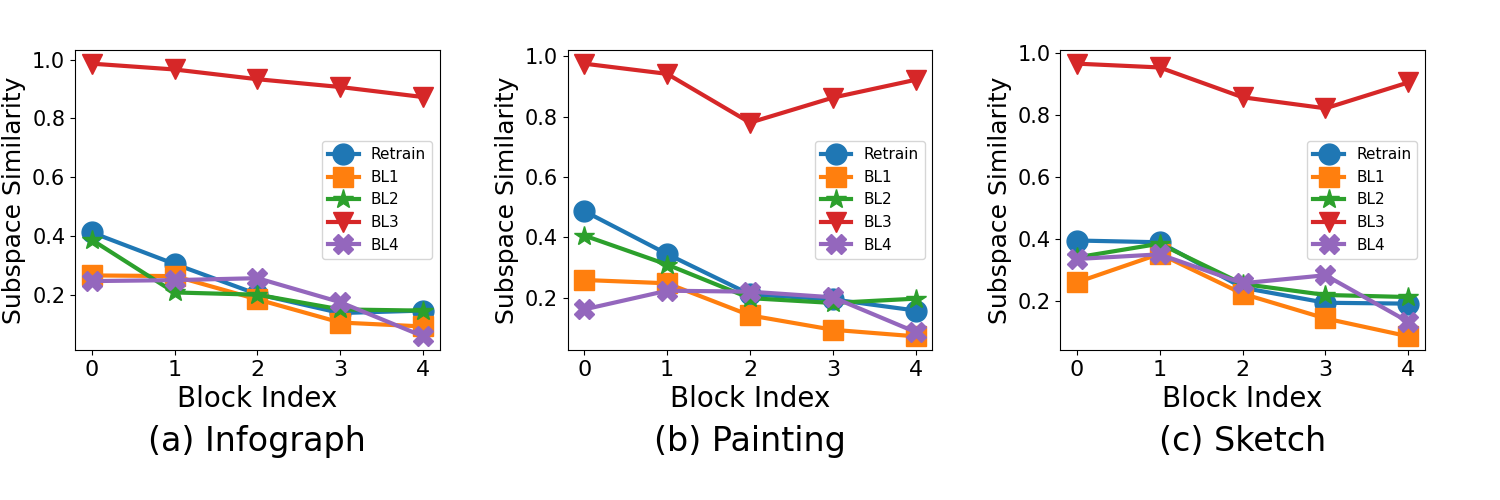}
\vspace{-25pt}
\caption{Comparative Analysis of Subspace Similarity in Feature Extractors Before and After Unlearning in the Target Domain of DomainNet.}
\label{fig_subspace}
\vspace{-20pt}
\end{figure}

As shown in Figure \ref{fig_subspace}, we display the subspace similarity of feature extractors before and after the application of various unlearning methods in the DomainNet. More results from other datasets are shown in supplementary materials. The results indicate that methods like Increase Loss maintain high subspace similarity across different layers, suggesting that features are extensively reused, which impedes effective unlearning. Conversely, methods that achieve better domain unlearning, such as FedEraser and Rapid Retraining, show a decrease in subspace similarity as layer depth increases. This decrease in similarity suggests that these methods are more effective in modifying or discarding the feature representations associated with the unlearned domain, thereby reducing the influence of the original domain-specific information in the higher layers of the network. This alteration in feature space demonstrates a more thorough unlearning process, aligning with the intended outcomes of these methods, which is to ensure that the domain knowledge is not merely suppressed but substantially erased from the model's memory.

\section{New validation methods for Federated domain unlearning}

\subsection{Current Verification Methods and Their Limitations}
In Federated Domain Unlearning, traditional verification methods such as Membership Inference Attacks \cite{shokri2017membership} (MIA) and Backdoor Attacks \cite{gu2017badnets,liu2018trojaning} face substantial challenges. MIAs, aimed at determining if a model retains information about specific domains post-unlearning, are hindered by the inherent heterogeneity of federated systems, where varied data distributions across nodes complicate the generalization of findings and the evaluation of unlearning efficacy. On the other hand, Backdoor Attacks, which test whether a model behaves normally under usual conditions but can be maliciously triggered by specific inputs, require detailed knowledge of each domain's characteristics to design effective triggers. This requirement not only increases the complexity of executing these attacks but also risks degrading the performance of the federated model by obliging it to learn both legitimate and malicious patterns \cite{huang2023rethinking}. Additionally, the security and compliance risks associated with deploying backdoor attacks in environments with sensitive data limit their practical applicability. Given these issues, while MIAs and Backdoor Attacks provide valuable insights into privacy and security vulnerabilities under standard conditions, their effectiveness diminishes in the complex and heterogeneous environment of federated domain unlearning \cite{huang2023rethinking,li2021fedbn}, highlighting the need for new strategies or adapted methods that address these unique challenges without compromising the model's integrity and performance.

\subsection{Methodology Overview}
Our proposed verification framework consists of four stages: domain representative sample selection, marker generation, marker injection, and verification. The key idea of our approach lies in creating a visually inconspicuous marker that is seamlessly integrated with the training of the target unlearning domain. This integration is designed to bind the marker indelibly into the domain-specific learning patterns of the model.

\subsubsection{Domain Representative Sample Selection} 
This initial phase involves the careful identification and selection of a representative set of samples that accurately encapsulates the full spectrum of features intrinsic to the target domain. The chosen samples must be reflective of the domain's diversity \cite{goodfellow2013empirical,toneva2018empirical}, ensuring that the generated marker is representative of the domain's entirety. 
To achieve this, we introduce a metric \( R_i \) for each sample \( x_i \), which aggregates the information pertaining to the sample's Learning and Forgetting events and reflects its importance to the domain learning process. For a given client domain, let \( \hat{y}_{i}^{(t)} \) be the predicted label for the sample \( x_i \) after \( t \) iterations of Stochastic Gradient Descent (SGD), with \( \hat{y}_{i}^{(t)} = \text{arg max}_k p(y_{ik} | x_i ; \theta^{(t)}) \). The binary variable \( \text{acc}_i^{(t)} = 1 \) if the prediction matches the true label \( y_i \), indicating correct classification at time step \( t \). We define a Forgetting event for sample \( x_i \) when there is a reduction in accuracy between consecutive iterations: \( \text{acc}_i^{(t)} > \text{acc}_i^{(t+1)} \). Conversely, a Learning event occurs when \( \text{acc}_i^{(t)} < \text{acc}_i^{(t+1)} \).

To calculate the representativeness metric \( R_i \), we consider the frequency and pattern of these events. Samples that undergo frequent Learning events, and few to no Forgetting events, are considered highly representative of the client domain. We can formally denote \( R_i \) as follows:
\begin{equation}
        R_i =  \sum_{t} (\text{Forgetting events}) - \sum_{t} (\text{Learning events})  
\end{equation}
When assessing sample representativeness for federated domain unlearning, special consideration is given to the role of "Unforgettable" samples \cite{toneva2018empirical}. These are defined as samples for which there exists a training step \( t^* \) such that for all subsequent steps \( k \geq t^* \), the accuracy \( \text{acc}_i^{(k)} = 1 \), meaning the sample continues to be correctly classified without error. Any sample \( x_i \) with a \( R_i \) that remains unchanged or their Forgetting events does not meet a minimal threshold increase \( \delta \) over the duration of the model's training is classified as Unforgettable and thus excluded from further analysis and processing in the marker generation and injection stages.
Samples with \( R_i \) exceeding a predefined threshold \( \tau \) are classified as representative samples of the client domain:
\begin{equation}
    x_i \text{ is representative if } R_i > \tau 
\end{equation}
Through this mechanism, we can systematically identify and select samples that are most indicative of the domain's characteristics. These representative samples are then used in the subsequent marker generation and injection phases, enabling precise verification of domain unlearning. 

\subsubsection{Marker Generation. } 
The process of marker generation is crucial as it facilitates the association of the marker with the domain's knowledge without being overtly detectable. Unlike traditional backdoor attacks \cite{gu2017badnets} where markers can be visually obvious and potentially degrade the model’s overall performance, our approach ensures that the injected marker remains subtle and does not disrupt the domain-specific learning processes.

To achieve this, we employ a model \( M_{\delta} \) designed to generate adversarial perturbations that act as markers. These markers are formulated to blend seamlessly with the domain representative samples, transforming them into modified samples that are nearly indistinguishable from their originals. However, these samples are configured to subtly mislead the federated model \( F_{\psi} \) into making incorrect predictions specifically for the targeted class. 
This strategic manipulation of \( F_{\psi} \), using local updates based on \( M_{\delta} \), allows us to adjust the adversarial generation adaptively, maintaining efficacy despite potential changes in the global model across different training rounds. The marker generation is modeled as an adversarial perturbation on the input, defined by the transformation:
\begin{equation}
    T_{\delta}(z) = z + M_{\delta}(z), \quad \|M_{\delta}(z)\|_{\infty} \leq \epsilon, \quad \forall z
\end{equation}
where $ \|\|_{\infty}$ is the infinity norm and parameter $\epsilon$ controls the visual stealthiness of the marker. \( M_{\delta} \) takes an input \( z \) and generates adversarial noise within the same input space, ensuring the stealthiness of the backdoor attack. To refine \( M_{\delta} \) effectively, we leverage the local model as a surrogate to update the attack model using the objective function:
\begin{equation}
    \delta \leftarrow \delta - \eta_{\delta} \sum L_{\delta}(h_{\theta}(T_{\delta}(z)), y_T), \quad z \in D
\end{equation}
In this equation, \( h_{\theta} \) denotes the locally trained model, and \( y_T \) represents the targeted label.

\subsubsection{Marker Injection.}  
In the Marker Injection phase, the objective is to seamlessly integrate the marker into the domain-specific learning process by training the local model \( h_{\theta} \) on both the domain-representative, clean samples \( z \) and the modified samples \( T_{\delta}(z) \) that include the marker. The goal is to maintain high feature similarity between the marked and unmarked samples, ensuring that the local model's performance remains robust on clean data while subtly shifting its response to the marked data toward a targeted outcome. This is achieved by optimizing the model with an objective function that not only minimizes the traditional loss between the predicted and actual labels of clean data but also aligns the features of clean and marked samples in terms of cosine similarity. The complete objective function is expressed as:
\begin{equation}
\begin{aligned}
     \min_{\theta} \Bigg( &\sum_{z \in D_R} \mu \cdot L(h_{\theta}(z), y) + 
     \lambda \cdot L(h_{\theta}(T_{\delta}(z)), y_T) + \\
     &  \text{Cos}(F(h_{\theta}(z)), F(h_{\theta}(T_{\delta}(z))) \Bigg)
\end{aligned}
\end{equation}
Here, \( L \) denotes the loss function guiding the model training, \( y \) and \( y_T \) are the true labels for clean and target classes, \( F(\cdot) \) represents the feature extraction process within the model, and \( \lambda \) and \( \mu \) are tunable parameters that balance the classification accuracy on the original data and data with the marker. This formulation ensures that the marker becomes an integral yet inconspicuous part of the learning process, effectively embedding within the domain data without compromising the model’s overall accuracy on unmarked samples.

\subsubsection{Verification.} In the Verification phase, we evaluate the effectiveness of unlearning by measuring the local model \( h_{\theta} \)'s accuracy on a test set where samples are embedded with the marker and have flipped labels to a specific target class \( y_T \). This setup tests whether the marker, intended to signal domain knowledge, has been adequately unlearned. High accuracy on these marked samples indicates that the model retains the marker-related knowledge, suggesting incomplete unlearning. Conversely, reduced accuracy implies successful unlearning, as the model no longer associates the marker with \( y_T \). This accuracy is calculated as:
\[ \text{Accuracy} = \frac{1}{|D_R|} \sum_{z \in D_R} \mathbb{I}[\hat{y}_T = y_T] \]
where \( D_R \) is the set of marked samples, \( \hat{y}_T \) is the predicted label, and \( \mathbb{I} \) is the indicator function. This verification is critical for assessing the integrity of the unlearning process and ensuring the model’s resilience against manipulations.

\vspace{-4pt}
\subsection{Validation Results}
\subsubsection{Experiment Settings.} We follow the setup of the experiment in Section 4.1 to verify the domain unlearning methods. The generation of markers is accomplished through the utilization of the U-Net architecture \cite{ronneberger2015u}. 
We set the hyperparameter $\tau = 10, \lambda=0.5$ and $\mu=0.5$ for all experiments. The marker generation is triggered at the last 10 rounds of the full training. We also compare the original backdoor attack, which introduces a `pixel
pattern' trigger of size 3x3 using the Adversarial Robust-ness Toolbox \cite{croce2020robustbench}.

\subsubsection{Efficiency Evaluation.} We show the efficacy of our verification method in terms of domain sensitivity and specificity in Domain-Digits datasets. Other datasets' results are shown in supplementary materials. The results, as detailed in Tables \ref{verifyresult1} and \ref{verifyresult2}, clearly demonstrate the advantages of our method over traditional backdoor attacks in evaluating the accuracy of unlearned domains with various unlearning techniques. Our method effectively eliminates background noise and accurately binds and assesses the targeted domain meant for unlearning, ensuring precise verification of the unlearning process. In contrast, backdoor attacks, hindered by the complexity of the domain, introduce extraneous noise that can impede accurate recognition. Additionally, our verification approach exerts minimal impact on the normal operational performance of the model. This is in stark contrast to backdoor methods, which, as observed in the domain-digital dataset, can decrease performance by as much as 13\%. This comparison underscores our method's superior capability to not only accurately assess unlearning but also preserve the integrity of the model's original training dynamics.

\subsubsection{Ablation Study.} 
\begin{figure}[!t]
\centering
\includegraphics[width=0.9\linewidth]{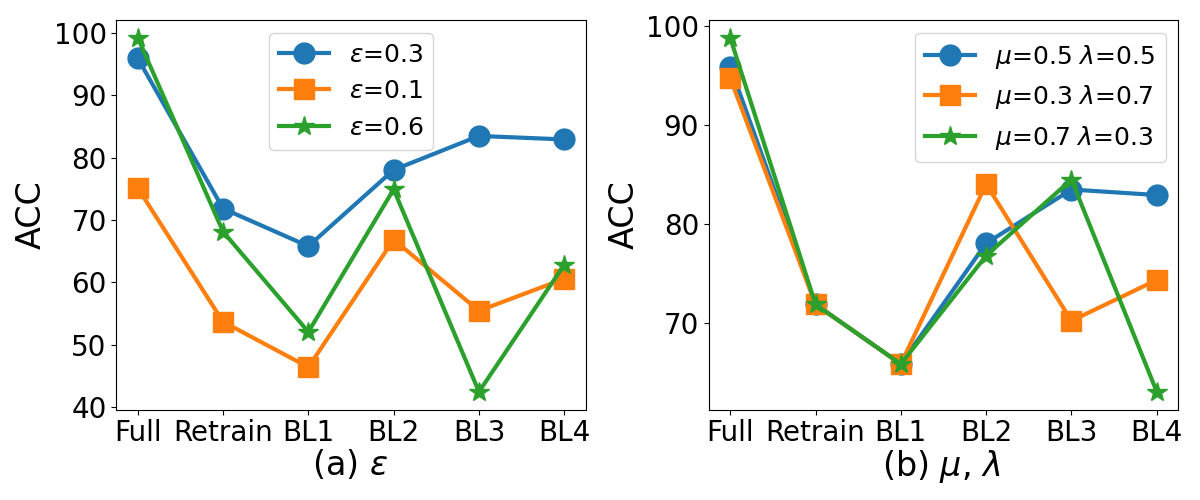}
\vspace{-15pt}
\caption{Validation Results of Our Verification Method Under Different Parameters.}
\label{fig_ablataion}
\vspace{-15pt}
\end{figure}

In our ablation study, we examined the impact of hyperparameters \(\epsilon\), \(\mu\), and \(\lambda\) on our verification method. The parameter \(\epsilon\) controls the visual stealthiness of the marker; findings presented in Figure \ref{fig_ablataion} demonstrate that a larger \(\epsilon\), leading to more conspicuous markers, simplifies the learning process for the generative model, whereas a smaller \(\epsilon\) complicates it by requiring fine-tuned detection reliant on the variance of the surrogate model across FL training rounds. Parameters \(\mu\) and \(\lambda\) balance the loss contributions from clean and marker-injected data; a higher \(\mu\) compared to \(\lambda\) shifts the model's focus towards optimizing performance on clean data, enabling rapid convergence to the performance of an unmodified classifier. This study emphasizes the need for precise calibration of these parameters to ensure balanced learning and effective unlearning verification.

\begin{table}[]

\caption{Evaluation Results of Our Verification Method and Backdoor Attacks on Original Model Performance in Domain-Digits dataset.}
\vspace{-10pt}
\label{verifyresult1}
\resizebox{\linewidth}{!}{

\begin{tabular}{ccccccc}
\hline
Domain                       & Method   & \multicolumn{5}{c}{\textbf{Test  ACC  For All  Domain}} \\ \hline
/                            & /        & MNIST   & SVHN    & USPS    & SynthDigits   & MNIST-M   \\  \hline
/                            & Clean    & 98.91   & 83.36   & 97.42   & 93.57         & 90.40     \\\hline
\multirow{2}{*}{MNIST}       & Backdoor & 98.69   & 82.84   & 96.83   & 93.48         & 89.44     \\ 
                             & Ours     & 98.91   & 82.30   & 97.90   & 94.73         & 92.72     \\\hline
\multirow{2}{*}{SVHN}        & Backdoor & 98.91   & 80.81   & 97.47   & 93.13         & 90.21     \\
                             & Ours     & 98.89   & 80.89   & 98.33   & 93.68         & 91.36     \\ \hline
\multirow{2}{*}{USPS}        & Backdoor & 98.41   & 79.79   & 95.32   & 91.94         & 87.30     \\
                             & Ours     & 98.85   & 85.16   & 96.72   & 94.80         & 92.16     \\ \hline
\multirow{2}{*}{SynthDigits} & Backdoor & 98.87   & 81.66   & 97.47   & 91.93         & 89.38     \\
                             & Ours     & 98.83   & 83.02   & 97.96   & 93.60         & 91.55     \\ \hline
\multirow{2}{*}{MNIST-M}     & Backdoor & 98.66   & 81.24   & 97.53   & 92.76         & 88.19     \\
                             & Ours     & 98.80   & 83.78   & 98.12   & 95.22         & 89.77    \\ \hline
\end{tabular}
}
\vspace{-10pt}
\end{table}




\begin{table}[!t]
\small
\caption{Evaluation of Our Verification Method for Domain Unlearning and Backdoor Attack Across Various Methods in Domain-Digits dataset.}
\label{verifyresult2}
\vspace{-10pt}
\resizebox{\linewidth}{!}{

\begin{tabular}{ccccccc}
\hline
\multirow{2}{*}{Doamin} & \multirow{2}{*}{Method} & \multicolumn{5}{c}{\multirow{2}{*}{\textbf{Train ACC  For BaseLine}}} \\
                             &          & \multicolumn{5}{c}{}                    \\ \hline
/                            & /        & Retrain & BL1   & BL2   & BL3   & BL4   \\ \hline
\multirow{3}{*}{MNIST}       & Clean    & 97.82   & 96.83 & 95.21 & 96.84 & 98.66 \\ 
                             & Backdoor & 0.39    & 0.45  & 0.20  & 0.42  & 0.48  \\
                             & Ours     & 0       & 0.66  & 10.26 & 42.60 & 19.62 \\ \hline
\multirow{3}{*}{SVHN}        & Clean    & 67.84   & 62.38 & 63.57 & 73.42 & 73.45 \\
                             & Backdoor & 1.29    & 2.75  & 1.38  & 42.64 & 8.49  \\
                             & Ours     & 0       & 6.03  & 6.19  & 11.63 & 11.07 \\ \hline
\multirow{3}{*}{USPS}        & Clean    & 89.33   & 88.94 & 87.95 & 82.83 & 91.87 \\
                             & Backdoor & 0.91    & 0.68  & 1.25  & 9.53  & 0.29  \\
                             & Ours     & 0       & 3.78  & 2.78  & 12.84 & 14.88 \\ \hline
\multirow{3}{*}{SynthDigits} & Clean    & 82.31   & 77.78 & 77.75 & 85.49 & 87.81 \\
                             & Backdoor & 0.56    & 1.16  & 2.45  & 24.33 & 2.46  \\
                             & Ours     & 0       & 2.34  & 2.74  & 9.52  & 2.19  \\ \hline
\multirow{3}{*}{MNIST-M}     & Clean    & 69.90   & 65.84 & 68.73 & 72.60 & 77.60 \\
                             & Backdoor & 5.51    & 6.40  & 8.76  & 20.71 & 5.27  \\
                             & Ours     & 0       & 4.14  & 3.82  & 29.05 & 4.24  \\ \hline
\end{tabular}}
\vspace{-10pt}
\end{table}

\section{Conclusion}

In our study, we investigate federated unlearning in multi-domain settings, highlighting major challenges in domain-specific unlearning, especially in preserving domain sensitivities and ensuring domain independence. We documented these complexities and identified persistent challenges and areas for improvement. Additionally, we introduced new verification methods to improve the robustness and effectiveness of unlearning in federated domains.

\clearpage

\bibliographystyle{ACM-Reference-Format}
\bibliography{sample-base}


\begin{thebibliography}{40}


\ifx \showCODEN    \undefined \def \showCODEN     #1{\unskip}     \fi
\ifx \showDOI      \undefined \def \showDOI       #1{#1}\fi
\ifx \showISBNx    \undefined \def \showISBNx     #1{\unskip}     \fi
\ifx \showISBNxiii \undefined \def \showISBNxiii  #1{\unskip}     \fi
\ifx \showISSN     \undefined \def \showISSN      #1{\unskip}     \fi
\ifx \showLCCN     \undefined \def \showLCCN      #1{\unskip}     \fi
\ifx \shownote     \undefined \def \shownote      #1{#1}          \fi
\ifx \showarticletitle \undefined \def \showarticletitle #1{#1}   \fi
\ifx \showURL      \undefined \def \showURL       {\relax}        \fi
\providecommand\bibfield[2]{#2}
\providecommand\bibinfo[2]{#2}
\providecommand\natexlab[1]{#1}
\providecommand\showeprint[2][]{arXiv:#2}

\bibitem[{California Department of Justice}(2020)]%
        {CCPA2020}
\bibfield{author}{\bibinfo{person}{{California Department of Justice}}.} \bibinfo{year}{2020}\natexlab{}.
\newblock \bibinfo{booktitle}{\emph{California Consumer Privacy Act (CCPA)}}.
\newblock
\urldef\tempurl%
\url{https://oag.ca.gov/privacy/ccpa}
\showURL{%
\tempurl}


\bibitem[Che et~al\mbox{.}(2023)]%
        {che2023fast}
\bibfield{author}{\bibinfo{person}{Tianshi Che}, \bibinfo{person}{Yang Zhou}, \bibinfo{person}{Zijie Zhang}, \bibinfo{person}{Lingjuan Lyu}, \bibinfo{person}{Ji Liu}, \bibinfo{person}{Da Yan}, \bibinfo{person}{Dejing Dou}, {and} \bibinfo{person}{Jun Huan}.} \bibinfo{year}{2023}\natexlab{}.
\newblock \showarticletitle{Fast federated machine unlearning with nonlinear functional theory}. In \bibinfo{booktitle}{\emph{International conference on machine learning}}. PMLR, \bibinfo{pages}{4241--4268}.
\newblock


\bibitem[Croce et~al\mbox{.}(2020)]%
        {croce2020robustbench}
\bibfield{author}{\bibinfo{person}{Francesco Croce}, \bibinfo{person}{Maksym Andriushchenko}, \bibinfo{person}{Vikash Sehwag}, \bibinfo{person}{Edoardo Debenedetti}, \bibinfo{person}{Nicolas Flammarion}, \bibinfo{person}{Mung Chiang}, \bibinfo{person}{Prateek Mittal}, {and} \bibinfo{person}{Matthias Hein}.} \bibinfo{year}{2020}\natexlab{}.
\newblock \showarticletitle{Robustbench: a standardized adversarial robustness benchmark}.
\newblock \bibinfo{journal}{\emph{arXiv preprint arXiv:2010.09670}} (\bibinfo{year}{2020}).
\newblock


\bibitem[{European Parliament and Council of the European Union}(2016)]%
        {GDPR2016}
\bibfield{author}{\bibinfo{person}{{European Parliament and Council of the European Union}}.} \bibinfo{year}{2016}\natexlab{}.
\newblock \bibinfo{booktitle}{\emph{Regulation (EU) 2016/679 (General Data Protection Regulation)}}.
\newblock
\urldef\tempurl%
\url{https://eur-lex.europa.eu/eli/reg/2016/679/oj}
\showURL{%
\tempurl}
\newblock
\shownote{Art. 17 - Right to erasure (’right to be forgotten’)}.


\bibitem[Ganin and Lempitsky(2015)]%
        {ganin2015unsupervised}
\bibfield{author}{\bibinfo{person}{Yaroslav Ganin} {and} \bibinfo{person}{Victor Lempitsky}.} \bibinfo{year}{2015}\natexlab{}.
\newblock \showarticletitle{Unsupervised domain adaptation by backpropagation}. In \bibinfo{booktitle}{\emph{International conference on machine learning}}. PMLR, \bibinfo{pages}{1180--1189}.
\newblock


\bibitem[Gao et~al\mbox{.}(2024)]%
        {gao2024verifi}
\bibfield{author}{\bibinfo{person}{Xiangshan Gao}, \bibinfo{person}{Xingjun Ma}, \bibinfo{person}{Jingyi Wang}, \bibinfo{person}{Youcheng Sun}, \bibinfo{person}{Bo Li}, \bibinfo{person}{Shouling Ji}, \bibinfo{person}{Peng Cheng}, {and} \bibinfo{person}{Jiming Chen}.} \bibinfo{year}{2024}\natexlab{}.
\newblock \showarticletitle{Verifi: Towards verifiable federated unlearning}.
\newblock \bibinfo{journal}{\emph{IEEE Transactions on Dependable and Secure Computing}} (\bibinfo{year}{2024}).
\newblock


\bibitem[Gong et~al\mbox{.}(2012)]%
        {gong2012geodesic}
\bibfield{author}{\bibinfo{person}{Boqing Gong}, \bibinfo{person}{Yuan Shi}, \bibinfo{person}{Fei Sha}, {and} \bibinfo{person}{Kristen Grauman}.} \bibinfo{year}{2012}\natexlab{}.
\newblock \showarticletitle{Geodesic flow kernel for unsupervised domain adaptation}. In \bibinfo{booktitle}{\emph{2012 IEEE conference on computer vision and pattern recognition}}. IEEE, \bibinfo{pages}{2066--2073}.
\newblock


\bibitem[Goodfellow et~al\mbox{.}(2013)]%
        {goodfellow2013empirical}
\bibfield{author}{\bibinfo{person}{Ian~J Goodfellow}, \bibinfo{person}{Mehdi Mirza}, \bibinfo{person}{Da Xiao}, \bibinfo{person}{Aaron Courville}, {and} \bibinfo{person}{Yoshua Bengio}.} \bibinfo{year}{2013}\natexlab{}.
\newblock \showarticletitle{An empirical investigation of catastrophic forgetting in gradient-based neural networks}.
\newblock \bibinfo{journal}{\emph{arXiv preprint arXiv:1312.6211}} (\bibinfo{year}{2013}).
\newblock


\bibitem[Gu et~al\mbox{.}(2017)]%
        {gu2017badnets}
\bibfield{author}{\bibinfo{person}{Tianyu Gu}, \bibinfo{person}{Brendan Dolan-Gavitt}, {and} \bibinfo{person}{Siddharth Garg}.} \bibinfo{year}{2017}\natexlab{}.
\newblock \showarticletitle{Badnets: Identifying vulnerabilities in the machine learning model supply chain}.
\newblock \bibinfo{journal}{\emph{arXiv preprint arXiv:1708.06733}} (\bibinfo{year}{2017}).
\newblock


\bibitem[Halimi et~al\mbox{.}(2022)]%
        {halimi2022federated}
\bibfield{author}{\bibinfo{person}{Anisa Halimi}, \bibinfo{person}{Swanand Kadhe}, \bibinfo{person}{Ambrish Rawat}, {and} \bibinfo{person}{Nathalie Baracaldo}.} \bibinfo{year}{2022}\natexlab{}.
\newblock \showarticletitle{Federated unlearning: How to efficiently erase a client in fl?}
\newblock \bibinfo{journal}{\emph{arXiv preprint arXiv:2207.05521}} (\bibinfo{year}{2022}).
\newblock


\bibitem[Huang et~al\mbox{.}(2022a)]%
        {huang2022learn}
\bibfield{author}{\bibinfo{person}{Wenke Huang}, \bibinfo{person}{Mang Ye}, {and} \bibinfo{person}{Bo Du}.} \bibinfo{year}{2022}\natexlab{a}.
\newblock \showarticletitle{Learn from others and be yourself in heterogeneous federated learning}. In \bibinfo{booktitle}{\emph{Proceedings of the IEEE/CVF Conference on Computer Vision and Pattern Recognition}}. \bibinfo{pages}{10143--10153}.
\newblock


\bibitem[Huang et~al\mbox{.}(2022b)]%
        {huang2022few}
\bibfield{author}{\bibinfo{person}{Wenke Huang}, \bibinfo{person}{Mang Ye}, \bibinfo{person}{Bo Du}, {and} \bibinfo{person}{Xiang Gao}.} \bibinfo{year}{2022}\natexlab{b}.
\newblock \showarticletitle{Few-shot model agnostic federated learning}. In \bibinfo{booktitle}{\emph{Proceedings of the 30th ACM International Conference on Multimedia}}. \bibinfo{pages}{7309--7316}.
\newblock


\bibitem[Huang et~al\mbox{.}(2023)]%
        {huang2023rethinking}
\bibfield{author}{\bibinfo{person}{Wenke Huang}, \bibinfo{person}{Mang Ye}, \bibinfo{person}{Zekun Shi}, \bibinfo{person}{He Li}, {and} \bibinfo{person}{Bo Du}.} \bibinfo{year}{2023}\natexlab{}.
\newblock \showarticletitle{Rethinking federated learning with domain shift: A prototype view}. In \bibinfo{booktitle}{\emph{2023 IEEE/CVF Conference on Computer Vision and Pattern Recognition (CVPR)}}. IEEE, \bibinfo{pages}{16312--16322}.
\newblock


\bibitem[Hull(1994)]%
        {hull1994database}
\bibfield{author}{\bibinfo{person}{Jonathan~J. Hull}.} \bibinfo{year}{1994}\natexlab{}.
\newblock \showarticletitle{A database for handwritten text recognition research}.
\newblock \bibinfo{journal}{\emph{IEEE Transactions on pattern analysis and machine intelligence}} \bibinfo{volume}{16}, \bibinfo{number}{5} (\bibinfo{year}{1994}), \bibinfo{pages}{550--554}.
\newblock


\bibitem[Jeong et~al\mbox{.}(2024)]%
        {jeong2024sok}
\bibfield{author}{\bibinfo{person}{Hyejun Jeong}, \bibinfo{person}{Shiqing Ma}, {and} \bibinfo{person}{Amir Houmansadr}.} \bibinfo{year}{2024}\natexlab{}.
\newblock \showarticletitle{SoK: Challenges and Opportunities in Federated Unlearning}.
\newblock \bibinfo{journal}{\emph{arXiv preprint arXiv:2403.02437}} (\bibinfo{year}{2024}).
\newblock


\bibitem[Kairouz et~al\mbox{.}(2019)]%
        {Kairouz2019}
\bibfield{author}{\bibinfo{person}{Peter Kairouz} {et~al\mbox{.}}} \bibinfo{year}{2019}\natexlab{}.
\newblock \showarticletitle{{Advances and Open Problems in Federated Learning}}.
\newblock \bibinfo{journal}{\emph{arXiv}} (\bibinfo{date}{dec} \bibinfo{year}{2019}).
\newblock
\showeprint{1912.04977}


\bibitem[Kalis(2014)]%
        {kalis2014google}
\bibfield{author}{\bibinfo{person}{Sarah~M Kalis}.} \bibinfo{year}{2014}\natexlab{}.
\newblock \showarticletitle{Google spain sl, google inc. v. agencia espanola de proteccion de datos, mario costeja gonzalez: An entitlement to erasure and its endlenss effects}.
\newblock \bibinfo{journal}{\emph{Tul. J. Int'l \& Comp. L.}}  \bibinfo{volume}{23} (\bibinfo{year}{2014}), \bibinfo{pages}{589}.
\newblock


\bibitem[Kone{\v{c}}n{\'{y}} et~al\mbox{.}(2016)]%
        {Konecny2016}
\bibfield{author}{\bibinfo{person}{Jakub Kone{\v{c}}n{\'{y}}}, \bibinfo{person}{H.~Brendan McMahan}, \bibinfo{person}{Daniel Ramage}, {and} \bibinfo{person}{Peter Richt{\'{a}}rik}.} \bibinfo{year}{2016}\natexlab{}.
\newblock \showarticletitle{{Federated Optimization: Distributed Machine Learning for On-Device Intelligence}}.
\newblock \bibinfo{journal}{\emph{arXiv}} (\bibinfo{date}{oct} \bibinfo{year}{2016}).
\newblock
\showeprint[arxiv]{1610.02527}


\bibitem[Kornblith et~al\mbox{.}(2019)]%
        {kornblith2019similarity}
\bibfield{author}{\bibinfo{person}{Simon Kornblith}, \bibinfo{person}{Mohammad Norouzi}, \bibinfo{person}{Honglak Lee}, {and} \bibinfo{person}{Geoffrey Hinton}.} \bibinfo{year}{2019}\natexlab{}.
\newblock \showarticletitle{Similarity of neural network representations revisited}. In \bibinfo{booktitle}{\emph{International conference on machine learning}}. PMLR, \bibinfo{pages}{3519--3529}.
\newblock


\bibitem[LeCun et~al\mbox{.}(1998)]%
        {lecun1998gradient}
\bibfield{author}{\bibinfo{person}{Yann LeCun}, \bibinfo{person}{L{\'e}on Bottou}, \bibinfo{person}{Yoshua Bengio}, {and} \bibinfo{person}{Patrick Haffner}.} \bibinfo{year}{1998}\natexlab{}.
\newblock \showarticletitle{Gradient-based learning applied to document recognition}.
\newblock \bibinfo{journal}{\emph{Proc. IEEE}} \bibinfo{volume}{86}, \bibinfo{number}{11} (\bibinfo{year}{1998}), \bibinfo{pages}{2278--2324}.
\newblock


\bibitem[Li et~al\mbox{.}(2020a)]%
        {Li2020_SPM}
\bibfield{author}{\bibinfo{person}{Tian Li}, \bibinfo{person}{Anit~Kumar Sahu}, \bibinfo{person}{Ameet Talwalkar}, {and} \bibinfo{person}{Virginia Smith}.} \bibinfo{year}{2020}\natexlab{a}.
\newblock \showarticletitle{{Federated Learning: Challenges, Methods, and Future Directions}}.
\newblock \bibinfo{journal}{\emph{IEEE Signal Processing Magazine}} \bibinfo{volume}{37}, \bibinfo{number}{3} (\bibinfo{date}{may} \bibinfo{year}{2020}), \bibinfo{pages}{50--60}.
\newblock


\bibitem[Li et~al\mbox{.}(2020b)]%
        {li2020federated}
\bibfield{author}{\bibinfo{person}{Tian Li}, \bibinfo{person}{Anit~Kumar Sahu}, \bibinfo{person}{Manzil Zaheer}, \bibinfo{person}{Maziar Sanjabi}, \bibinfo{person}{Ameet Talwalkar}, {and} \bibinfo{person}{Virginia Smith}.} \bibinfo{year}{2020}\natexlab{b}.
\newblock \showarticletitle{Federated optimization in heterogeneous networks}.
\newblock \bibinfo{journal}{\emph{Proceedings of Machine learning and systems}}  \bibinfo{volume}{2} (\bibinfo{year}{2020}), \bibinfo{pages}{429--450}.
\newblock


\bibitem[Li et~al\mbox{.}(2021)]%
        {li2021fedbn}
\bibfield{author}{\bibinfo{person}{Xiaoxiao Li}, \bibinfo{person}{Meirui Jiang}, \bibinfo{person}{Xiaofei Zhang}, \bibinfo{person}{Michael Kamp}, {and} \bibinfo{person}{Qi Dou}.} \bibinfo{year}{2021}\natexlab{}.
\newblock \showarticletitle{Fedbn: Federated learning on non-iid features via local batch normalization}.
\newblock \bibinfo{journal}{\emph{arXiv preprint arXiv:2102.07623}} (\bibinfo{year}{2021}).
\newblock


\bibitem[Liu et~al\mbox{.}(2021)]%
        {liu2021federaser}
\bibfield{author}{\bibinfo{person}{Gaoyang Liu}, \bibinfo{person}{Xiaoqiang Ma}, \bibinfo{person}{Yang Yang}, \bibinfo{person}{Chen Wang}, {and} \bibinfo{person}{Jiangchuan Liu}.} \bibinfo{year}{2021}\natexlab{}.
\newblock \showarticletitle{Federaser: Enabling efficient client-level data removal from federated learning models}. In \bibinfo{booktitle}{\emph{2021 IEEE/ACM 29th International Symposium on Quality of Service (IWQOS)}}. IEEE, \bibinfo{pages}{1--10}.
\newblock


\bibitem[Liu et~al\mbox{.}(2018)]%
        {liu2018trojaning}
\bibfield{author}{\bibinfo{person}{Yingqi Liu}, \bibinfo{person}{Shiqing Ma}, \bibinfo{person}{Yousra Aafer}, \bibinfo{person}{Wen-Chuan Lee}, \bibinfo{person}{Juan Zhai}, \bibinfo{person}{Weihang Wang}, {and} \bibinfo{person}{Xiangyu Zhang}.} \bibinfo{year}{2018}\natexlab{}.
\newblock \showarticletitle{Trojaning attack on neural networks}. In \bibinfo{booktitle}{\emph{25th Annual Network And Distributed System Security Symposium (NDSS 2018)}}. Internet Soc.
\newblock


\bibitem[Liu et~al\mbox{.}(2022)]%
        {liu2022right}
\bibfield{author}{\bibinfo{person}{Yi Liu}, \bibinfo{person}{Lei Xu}, \bibinfo{person}{Xingliang Yuan}, \bibinfo{person}{Cong Wang}, {and} \bibinfo{person}{Bo Li}.} \bibinfo{year}{2022}\natexlab{}.
\newblock \showarticletitle{The right to be forgotten in federated learning: An efficient realization with rapid retraining}. In \bibinfo{booktitle}{\emph{IEEE INFOCOM 2022-IEEE Conference on Computer Communications}}. IEEE, \bibinfo{pages}{1749--1758}.
\newblock


\bibitem[Liu et~al\mbox{.}(2023)]%
        {liu2023survey}
\bibfield{author}{\bibinfo{person}{Ziyao Liu}, \bibinfo{person}{Yu Jiang}, \bibinfo{person}{Jiyuan Shen}, \bibinfo{person}{Minyi Peng}, \bibinfo{person}{Kwok-Yan Lam}, {and} \bibinfo{person}{Xingliang Yuan}.} \bibinfo{year}{2023}\natexlab{}.
\newblock \showarticletitle{A survey on federated unlearning: Challenges, methods, and future directions}.
\newblock \bibinfo{journal}{\emph{arXiv preprint arXiv:2310.20448}} (\bibinfo{year}{2023}).
\newblock


\bibitem[McMahan et~al\mbox{.}(2017)]%
        {mcmahan2017communication}
\bibfield{author}{\bibinfo{person}{Brendan McMahan}, \bibinfo{person}{Eider Moore}, \bibinfo{person}{Daniel Ramage}, \bibinfo{person}{Seth Hampson}, {and} \bibinfo{person}{Blaise~Aguera y Arcas}.} \bibinfo{year}{2017}\natexlab{}.
\newblock \showarticletitle{Communication-efficient learning of deep networks from decentralized data}. In \bibinfo{booktitle}{\emph{Artificial intelligence and statistics}}. PMLR, \bibinfo{pages}{1273--1282}.
\newblock


\bibitem[Netzer et~al\mbox{.}(2011)]%
        {netzer2011reading}
\bibfield{author}{\bibinfo{person}{Yuval Netzer}, \bibinfo{person}{Tao Wang}, \bibinfo{person}{Adam Coates}, \bibinfo{person}{Alessandro Bissacco}, \bibinfo{person}{Baolin Wu}, \bibinfo{person}{Andrew~Y Ng}, {et~al\mbox{.}}} \bibinfo{year}{2011}\natexlab{}.
\newblock \showarticletitle{Reading digits in natural images with unsupervised feature learning}. In \bibinfo{booktitle}{\emph{NIPS workshop on deep learning and unsupervised feature learning}}, Vol.~\bibinfo{volume}{2011}. Granada, Spain, \bibinfo{pages}{7}.
\newblock


\bibitem[Peng et~al\mbox{.}(2019)]%
        {peng2019moment}
\bibfield{author}{\bibinfo{person}{Xingchao Peng}, \bibinfo{person}{Qinxun Bai}, \bibinfo{person}{Xide Xia}, \bibinfo{person}{Zijun Huang}, \bibinfo{person}{Kate Saenko}, {and} \bibinfo{person}{Bo Wang}.} \bibinfo{year}{2019}\natexlab{}.
\newblock \showarticletitle{Moment matching for multi-source domain adaptation}. In \bibinfo{booktitle}{\emph{Proceedings of the IEEE/CVF international conference on computer vision}}. \bibinfo{pages}{1406--1415}.
\newblock


\bibitem[Ronneberger et~al\mbox{.}(2015)]%
        {ronneberger2015u}
\bibfield{author}{\bibinfo{person}{Olaf Ronneberger}, \bibinfo{person}{Philipp Fischer}, {and} \bibinfo{person}{Thomas Brox}.} \bibinfo{year}{2015}\natexlab{}.
\newblock \showarticletitle{U-net: Convolutional networks for biomedical image segmentation}. In \bibinfo{booktitle}{\emph{Medical image computing and computer-assisted intervention--MICCAI 2015: 18th international conference, Munich, Germany, October 5-9, 2015, proceedings, part III 18}}. Springer, \bibinfo{pages}{234--241}.
\newblock


\bibitem[Roy et~al\mbox{.}(2018)]%
        {roy2018effects}
\bibfield{author}{\bibinfo{person}{Prasun Roy}, \bibinfo{person}{Subhankar Ghosh}, \bibinfo{person}{Saumik Bhattacharya}, {and} \bibinfo{person}{Umapada Pal}.} \bibinfo{year}{2018}\natexlab{}.
\newblock \showarticletitle{Effects of degradations on deep neural network architectures}.
\newblock \bibinfo{journal}{\emph{arXiv preprint arXiv:1807.10108}} (\bibinfo{year}{2018}).
\newblock


\bibitem[Shokri et~al\mbox{.}(2017)]%
        {shokri2017membership}
\bibfield{author}{\bibinfo{person}{Reza Shokri}, \bibinfo{person}{Marco Stronati}, \bibinfo{person}{Congzheng Song}, {and} \bibinfo{person}{Vitaly Shmatikov}.} \bibinfo{year}{2017}\natexlab{}.
\newblock \showarticletitle{Membership inference attacks against machine learning models}. In \bibinfo{booktitle}{\emph{2017 IEEE symposium on security and privacy (SP)}}. IEEE, \bibinfo{pages}{3--18}.
\newblock


\bibitem[Tam et~al\mbox{.}(2023a)]%
        {tam2023federated}
\bibfield{author}{\bibinfo{person}{Kahou Tam}, \bibinfo{person}{Li Li}, \bibinfo{person}{Bo Han}, \bibinfo{person}{Chengzhong Xu}, {and} \bibinfo{person}{Huazhu Fu}.} \bibinfo{year}{2023}\natexlab{a}.
\newblock \showarticletitle{Federated noisy client learning}.
\newblock \bibinfo{journal}{\emph{IEEE Transactions on Neural Networks and Learning Systems}} (\bibinfo{year}{2023}).
\newblock


\bibitem[Tam et~al\mbox{.}(2023b)]%
        {tam2023fedcoop}
\bibfield{author}{\bibinfo{person}{Kahou Tam}, \bibinfo{person}{Li Li}, \bibinfo{person}{Yan Zhao}, {and} \bibinfo{person}{Chengzhong Xu}.} \bibinfo{year}{2023}\natexlab{b}.
\newblock \showarticletitle{FedCoop: Cooperative Federated Learning for Noisy Labels}.
\newblock In \bibinfo{booktitle}{\emph{ECAI 2023}}. \bibinfo{publisher}{IOS Press}, \bibinfo{pages}{2298--2306}.
\newblock


\bibitem[Toneva et~al\mbox{.}(2018)]%
        {toneva2018empirical}
\bibfield{author}{\bibinfo{person}{Mariya Toneva}, \bibinfo{person}{Alessandro Sordoni}, \bibinfo{person}{Remi Tachet~des Combes}, \bibinfo{person}{Adam Trischler}, \bibinfo{person}{Yoshua Bengio}, {and} \bibinfo{person}{Geoffrey~J Gordon}.} \bibinfo{year}{2018}\natexlab{}.
\newblock \showarticletitle{An empirical study of example forgetting during deep neural network learning}.
\newblock \bibinfo{journal}{\emph{arXiv preprint arXiv:1812.05159}} (\bibinfo{year}{2018}).
\newblock


\bibitem[Wang et~al\mbox{.}(2022)]%
        {wang2022federated}
\bibfield{author}{\bibinfo{person}{Junxiao Wang}, \bibinfo{person}{Song Guo}, \bibinfo{person}{Xin Xie}, {and} \bibinfo{person}{Heng Qi}.} \bibinfo{year}{2022}\natexlab{}.
\newblock \showarticletitle{Federated unlearning via class-discriminative pruning}. In \bibinfo{booktitle}{\emph{Proceedings of the ACM Web Conference 2022}}. \bibinfo{pages}{622--632}.
\newblock


\bibitem[Wu et~al\mbox{.}(2022b)]%
        {wu2022federated2}
\bibfield{author}{\bibinfo{person}{Chen Wu}, \bibinfo{person}{Sencun Zhu}, {and} \bibinfo{person}{Prasenjit Mitra}.} \bibinfo{year}{2022}\natexlab{b}.
\newblock \showarticletitle{Federated unlearning with knowledge distillation}.
\newblock \bibinfo{journal}{\emph{arXiv preprint arXiv:2201.09441}} (\bibinfo{year}{2022}).
\newblock


\bibitem[Wu et~al\mbox{.}(2022a)]%
        {wu2022federated}
\bibfield{author}{\bibinfo{person}{Leijie Wu}, \bibinfo{person}{Song Guo}, \bibinfo{person}{Junxiao Wang}, \bibinfo{person}{Zicong Hong}, \bibinfo{person}{Jie Zhang}, {and} \bibinfo{person}{Yaohong Ding}.} \bibinfo{year}{2022}\natexlab{a}.
\newblock \showarticletitle{Federated unlearning: Guarantee the right of clients to forget}.
\newblock \bibinfo{journal}{\emph{IEEE Network}} \bibinfo{volume}{36}, \bibinfo{number}{5} (\bibinfo{year}{2022}), \bibinfo{pages}{129--135}.
\newblock


\bibitem[Zhang et~al\mbox{.}(2023)]%
        {zhang2023federated}
\bibfield{author}{\bibinfo{person}{Ruipeng Zhang}, \bibinfo{person}{Qinwei Xu}, \bibinfo{person}{Jiangchao Yao}, \bibinfo{person}{Ya Zhang}, \bibinfo{person}{Qi Tian}, {and} \bibinfo{person}{Yanfeng Wang}.} \bibinfo{year}{2023}\natexlab{}.
\newblock \showarticletitle{Federated domain generalization with generalization adjustment}. In \bibinfo{booktitle}{\emph{Proceedings of the IEEE/CVF Conference on Computer Vision and Pattern Recognition}}. \bibinfo{pages}{3954--3963}.
\newblock


\end{thebibliography}

\clearpage

\section*{APPENDIX\\}
\appendices

\section{Implementation Details}

We conduct all our experiments using PyTorch. For Federated Prototypes Learning (FPL) \cite{huang2023rethinking}, Rapid Retraining \cite{liu2022right}, FedEraser \cite{liu2021federaser}, and Increase Loss \cite{halimi2022federated}, we utilize the authors' open-source code. We have re-implemented and adapted Class-Discriminative Pruning \cite{wang2022federated} to enable complete forgetting of an entire client. All experiments employ the cross-entropy loss function and use the SGD optimizer with a learning rate of 0.01 and a momentum of 0.9 across all datasets.

In our federated learning setup, we assign an entire domain of data to each client for each dataset. The experiments are conducted over 10 local update epochs and 50 global training rounds. The local batch size for all experiments is set to 64. We adhere to the hyper-parameters specified in the original work for FPL.

For the various methods employed in federated unlearning:
\begin{itemize}
    \item FedEraser is configured with a calibration ratio \( r=0.5 \) and a retaining interval \( \Delta t=1 \).
    \item Increase Loss sets an early stopping threshold \( \tau \) at 5, 20, and 20 for all experiments.
    \item The threshold \( R \) for Class-Discriminative Pruning is set to 0.7, aiming to ensure a high degree of specificity in pruning while maintaining overall network integrity.
\end{itemize}

\section{Experiment Details}

\subsection{Effectiveness of Existing Methods in Federated Domain Unlearning}

\begin{table*}[h]

\caption{
Evaluation of federated domain unlearning across various methods on Domain-Digital dataset.}

\label{result_digtis_acc}
\renewcommand{\arraystretch}{1.2}
\resizebox{\linewidth}{!}{%

\begin{tabular}{clcccccccccc}
\hline
\multicolumn{2}{c}{\multirow{2}{*}{Domain-Digits}} & \multicolumn{5}{c}{\multirow{2}{*}{Train  Accuracy  For Unlearn  Domain}} & \multicolumn{5}{c}{\multirow{2}{*}{Test  Accuracy  For All  Domain}} \\ 
\multicolumn{2}{c}{}                               & \multicolumn{5}{c}{}                                                 & \multicolumn{5}{c}{}                                            \\ \hline
Unlearn Domain                & BaseLine           & MNIST       & SVHN        & USPS        & SynthDigits  & MNIST-M     & MNIST      & SVHN       & USPS       & SynthDigits & MNIST-M    \\ \hline
/                             & Full learn         & 99.99±0.01  & 94.15±0.07  & 98.62±0.03  & 98.67±0.06   & 98.82±0.08  & 98.91±0.06 & 83.36±0.11 & 97.42±0.10 & 93.57±0.11  & 90.40±0.12 \\ \hline
\multirow{5}{*}{MNIST}        & Retrain            & 97.82±0.08  & 97.82±0.11  & 99.09±0.03  & 99.74±0.07   & 98.66±0.17  & 97.83±0.21 & 85.20±0.21 & 97.42±0.02 & 94.78±0.08  & 89.35±0.14 \\
                              & BL1 Repaid Retrain & 96.83±0.02  & 92.82±0.13  & 99.50±0.04  & 99.69±0.06   & 95.91±0.12  & 96.80±0.01 & 80.30±0.08 & 97.90±0.05 & 92.64±0.11  & 82.14±0.19 \\
                              & BL2 FedEraser      & 95.21±0.12  & 81.04±0.23  & 95.40±0.09  & 90.97±0.15   & 83.52±0.23  & 95.08±0.23 & 76.96±0.40 & 95.22±0.27 & 88.38±0.11  & 80.57±0.12 \\
                              & BL3 Increase Loss  & 96.84±0.09  & 97.06±0.17  & 99.41±0.04  & 99.93±0.03   & 99.61±0.11  & 95.96±0.11 & 84.32±0.05 & 97.8±0.02  & 94.46±0.04  & 90.19±0.10 \\
                              & BL4 Class Pruning  & 98.66±0.01  & 97.15±0.50  & 99.45±0.01  & 99.96±0.00   & 99.81±0.12  & 98.14±0.02 & 84.48±0.08 & 97.8±0.05  & 94.66±0.01  & 90.46±0.35 \\ \hline
\multirow{5}{*}{SVHN}         & Retrain            & 100.0±0.00  & 67.84±0.30  & 98.83±0.10  & 98.60±0.48   & 99.26±0.20  & 99.09±0.05 & 67.48±0.58 & 97.63±0.11 & 91.49±0.56  & 92.28±0.28 \\
                              & BL1 Repaid Retrain & 100.0±0.00  & 62.38±0.26  & 98.92±0.03  & 97.69±0.51   & 95.94±0.56  & 98.79±0.00 & 62.55±0.74 & 97.42±0.00 & 89.07±0.22  & 85.76±0.17 \\
                              & BL2 FedEraser      & 99.95±0.20  & 63.57±0.54  & 98.57±0.27  & 95.86±0.63   & 97.45±0.41  & 98.94±0.33 & 63.41±0.60 & 97.26±0.21 & 89.42±0.43  & 90.54±0.45 \\
                              & BL3 Increase Loss  & 99.97±0.02  & 73.42±0.48  & 99.37±0.02  & 99.72±0.11   & 99.78±0.03  & 98.99±0.03 & 70.12±0.58 & 98.01±0.05 & 93.59±0.09  & 92.36±0.14 \\
                              & BL4 Class Pruning  & 99.99±0.00  & 73.45±0.39  & 99.26±0.05  & 99.09±0.10   & 99.92±0.01  & 98.99±0.00 & 70.82±0.97 & 97.85±0.02 & 92.54±0.37  & 92.65±0.19 \\ \hline
\multirow{5}{*}{USPS}         & Retrain            & 99.89±0.01  & 93.92±0.24  & 89.33±0.01  & 99.54±0.25   & 99.60±0.12  & 98.49±0.07 & 83.35±0.01 & 89.30±0.03 & 94.05±0.17  & 91.21±0.15 \\
                              & BL1 Repaid Retrain & 99.91±0.00  & 87.24±0.28  & 88.94±0.02  & 99.10±0.09   & 98.82±0.16  & 98.49±0.12 & 78.73±0.04 & 88.87±0.15 & 91.68±0.02  & 86.89±0.04 \\
                              & BL2 FedEraser      & 98.37±0.10  & 79.35±0.19  & 87.95±0.25  & 89.93±0.27   & 89.10±0.18  & 97.64±0.15 & 75.94±021  & 86.88±0.11 & 87.62±0.21  & 85.56±0.15 \\
                              & BL3 Increase Loss  & 99.88±0.00  & 95.68±0.03  & 82.83±0.04  & 99.78±0.01   & 99.65±0.01  & 98.51±0.21 & 83.93±0.11 & 82.80±0.05 & 94.24±0.03  & 90.79±0.06 \\
                              & BL4 Class Pruning  & 99.93±0.00  & 95.7±0.02   & 91.87±0.02  & 99.93±0.06   & 99.87±0.02  & 98.74±0.05 & 84.40±0.02 & 91.83±0.06 & 94.55±0.11  & 91.58±0.08 \\ \hline
\multirow{5}{*}{SynthDigits}  & Retrain            & 99.96±0.00  & 87.21±0.34  & 99.31±0.27  & 82.31±0.61   & 99.33±0.02  & 98.90±0.02 & 76.51±0.59 & 97.31±0.17 & 82.50±0.65  & 91.54±0.18 \\
                              & BL1 Repaid Retrain & 99.97±0.00  & 80.22±0.79  & 98.39±0.10  & 77.78±0.27   & 97.88±0.32  & 98.64±0.09 & 71.22±0.82 & 96.88±0.17 & 77.98±0.12  & 87.14±0.27 \\
                              & BL2 FedEraser      & 99.26±0.02  & 77.36±0.71  & 95.66±0.23  & 77.75±0.27   & 93.22±0.30  & 98.14±0.01 & 71.72±0.81 & 94.57±0.42 & 78.15±0.32  & 87.49±0.13 \\
                              & BL3 Increase Loss  & 100.0±0.00  & 91.41±0.68  & 99.22±0.03  & 85.49±0.11   & 99.76±0.02  & 98.82±0.05 & 79.04±0.27 & 97.69±0.06 & 84.3±0.17   & 91.77±0.05 \\
                              & BL4 Class Pruning  & 100.0±0.00  & 93.12±0.12  & 99.58±0.04  & 87.81±0.05   & 99.89±0.03  & 98.96±0.00 & 80.59±0.10 & 98.06±0.05 & 86.88±0.16  & 91.59±0.19 \\ \hline
\multirow{5}{*}{MNIST-M}      & Retrain            & 99.87±0.00  & 94.94±0.85  & 99.57±0.01  & 99.78±0.07   & 69.9±0.09   & 98.39±0.02 & 84.21±0.62 & 98.49±0.05 & 94.73±0.09  & 70.35±0.10 \\
                              & BL1 Repaid Retrain & 99.72±0.01  & 89.14±0.39  & 99.39±0.01  & 99.37±0.02   & 65.84±0.11  & 97.90±0.02 & 80.40±0.11 & 97.58±0.00 & 93.13±0.02  & 65.49±0.07 \\
                              & BL2 FedEraser      & 99.50±0.22  & 92.22±1.30  & 99.26±0.21  & 99.13±0.64   & 68.73±0.25  & 97.75±0.09 & 82.79±0.71 & 98.12±0.58 & 94.16±0.21  & 68.76±0.39 \\
                              & BL3 Increase Loss  & 99.30±0.03  & 96.88±0.04  & 99.56±0.09  & 99.91±0.00   & 72.60±0.04  & 97.48±0.04 & 84.30±0.02 & 98.28±0.05 & 95.12±0.02  & 70.31±0.03 \\
                              & BL4 Class Pruning  & 99.92±0.00  & 96.57±0.03  & 99.84±0.00  & 99.95±0.00   & 77.60±0.01  & 98.55±0.02 & 84.84±0.10 & 98.23±0.07 & 95.33±0.17  & 75.22±0.21 \\ \hline
\end{tabular}
}
\end{table*}

\begin{table*}[!t]

\caption{
Evaluation of federated domain unlearning across various methods on Office-Caltech10 dataset.}

\label{result_office_acc}
\renewcommand{\arraystretch}{1.2}
\resizebox{\linewidth}{!}{%

\begin{tabular}{clcccccccc}
\hline
\multicolumn{2}{c}{Office-Caltech10}          & \multicolumn{4}{c}{Train  Accuracy  For All  Domain}                                       & \multicolumn{4}{c}{Test  Accuracy  For All  Domain}                                        \\ \hline
Unlearn Domain           & BaseLine           & Amazon              & Caltech             & Dslr                & Webcam              & Amazon              & Caltech             & Dslr                & Webcam              \\ \hline
/                        & Full learn         & 82.04±0.52          & 99.02±0.11          & 88.16±0.60          & 91.86±062           & 78.12±0.27          & 76±0.23             & 90.62±0.56          & 89.83±0.15          \\ \hline
\multirow{5}{*}{Amazon}  & Retrain            & \textbf{61.98±0.95} & 94.10±1.50          & 87.52±1.19          & 92.46±1.42          & \textbf{64.38±1.63} & 70.67±1.61          & 86.25±2.08          & 88.81±2.32          \\
                         & BL1 Repaid Retrain & \textbf{43.19±1.18} & 63.10±1.47          & 87.68±2.06          & 99.41±0.43          & \textbf{46.25±2.90} & 52.27±1.55          & 82.50±2.75          & 94.92±0.86          \\
                         & BL2 FedEraser      & \textbf{51.96±2.38} & 76.87±2.00          & 82.20±2.17          & 85.81±2.63          & \textbf{53.26±2.72} & 59.22±1.61          & 83.59±2.12          & 83.05±1.20          \\
                         & BL3 Increase Loss  & \textbf{71.04±1.15} & 98.95±0.53          & 89.12±1.65          & 95.42±0.82          & \textbf{73.23±0.42} & 74.49±0.72          & 88.75±1.50          & 89.83±1.07          \\
                         & BL4 Class Pruning  & \textbf{67.23±1.20} & 95.46±0.74          & 92.00±2.15          & 92.20±1.91          & \textbf{67.40±1.50} & 68.80±1.10          & 90.00±2.15          & 87.12±1.54          \\ \hline
\multirow{5}{*}{Caltehc} & Retrain            & 40.23±0.69          & \textbf{33.32±1.40} & 75.68±1.18          & 96.69±0.98          & 35.42±1.19          & \textbf{34.13±1.41} & 75.00±0.78          & 91.86±1.29          \\
                         & BL1 Repid Retrain  & 38.02±1.65          & \textbf{30.94±0.57} & 71.68±1.89          & 98.47±0.91          & 35.62±1.53          & \textbf{32.09±1.17} & 70.62±2.74          & 92.20±1.73          \\
                         & BL2 FedEraser      & 69.45±2.68          & \textbf{37.31±1.34} & 59.60±0.40          & 83.26±2.75          & 57.03±2.59          & \textbf{37.56±1.33} & 65.62±2.12          & 84.75±1.68          \\
                         & BL3 Increase Loss  & 87.96±0.32          & \textbf{91.85±0.34} & 81.76±1.06          & 91.02±0.42          & 80.73±0.66          & \textbf{69.87±0.33} & 76.88±1.53          & 86.78±0.68          \\
                         & BL4 Class Pruning  & 59.45±2.32          & \textbf{43.16±1.15} & 81.28±2.25          & 98.56±0.95          & 49.27±2.26          & \textbf{47.02±3.10} & 78.12±2.59          & 94.92±1.40          \\ \hline
\multirow{5}{*}{Dslr}    & Retrain            & 87.36±1.86          & 98.73±1.43          & \textbf{77.28±1.18} & 92.29±0.63          & 81.04±1.11          & 74.31±1.29          & \textbf{76.88±2.55} & 89.49±0.71          \\
                         & BL1 Repid Retrain  & 80.55±2.16          & 82.36±1.85          & \textbf{70.88±2.93} & 90.68±2.84          & 74.58±1.45          & 62.04±1.88          & \textbf{72.50±1.65} & 85.08±1.92          \\
                         & BL2 FedEraser      & 80.22±2.92          & 94.13±2.22          & \textbf{68.20±1.57} & 81.57±2.88          & 74.35±1.97          & 70.44±2.36          & \textbf{66.41±2.08} & 75.42±1.62          \\
                         & BL3 Increase Loss  & 89.19±2.01          & 98.82±1.09          & \textbf{80.64±1.20} & 90.00±1.11          & 82.40±1.52          & 74.22±1.01          & \textbf{80.62±1.67} & 82.03±0.96          \\
                         & BL4 Class Pruning  & 90.37±2.08          & 99.53±0.45          & \textbf{79.36±1.85} & 93.14±3.00          & 82.60±1.88          & 75.56±1.12          & \textbf{80.00±1.75} & 88.14±2.01          \\ \hline
\multirow{5}{*}{Webcam}  & Retrain            & 79.58±0.68          & 96.21±1.02          & 75.36±0.78          & \textbf{63.31±1.05} & 78.65±1.10          & 74.78±1.38          & 79.69±1.18          & \textbf{69.92±0.89} \\
                         & BL1 Repid Retrain  & 72.43±1.63          & 80.20±1.35          & 76.48±1.98          & \textbf{61.44±1.72} & 70.73±1.43          & 62.40±1.24          & 77.50±1.64          & \textbf{71.19±2.22} \\
                         & BL2 FedEraser      & 80.25±2.17          & 87.28±2.38          & 66.60±1.68          & \textbf{56.78±2.53} & 76.04±1.01          & 69.33±1.89          & 69.53±2.56          & \textbf{55.51±1.51} \\
                         & BL3 Increase Loss  & 87.91±1.40          & 98.80±0.57          & 75.20±1.69          & \textbf{65.25±0.97} & 82.92±1.37          & 74.13±1.67          & 81.88±1.25          & \textbf{69.83±2.25} \\
                         & BL4 Class Pruning  & 82.56±2.24          & 98.51±0.32          & 77.76±1.31          & \textbf{63.14±1.33} & 79.27±1.01          & 74.93±1.39          & 81.88±1.34          & \textbf{73.22±2.92} \\\hline
\end{tabular}
}
\end{table*}
We perform an empirical evaluation to determine the effectiveness of contemporary unlearning methods in various domains. The accuracy results for the unlearned domain and the remaining test accuracies for Domain-Digital and Office-Caltech10 are shown in Tables \ref{result_digtis_acc} and \ref{result_office_acc}. These experimental outcomes mirror those found in DomainNet. In summary, the present methods for federated unlearning introduce substantial challenges within the sphere of federated domain unlearning. These methods either compromise the learning of original domains while attempting to unlearn targeted domains or fail to completely remove the data of targeted domains. This dichotomy exposes a core limitation in existing methods, where the trade-off between effectively unlearning specific domain data and maintaining the integrity and performance of non-targeted domains is yet unresolved. The inability to selectively forget without residual effects calls for the development of more advanced techniques that can handle domain-specific unlearning without undermining the overall system's effectiveness and robustness.

\begin{table}[]

\caption{
CKA for three convolution layer and three fully connected layer of federated domain unlearning across various methods on Domain-Digital dataset.}

\label{cka_digit}
\renewcommand{\arraystretch}{1.2}
\resizebox{\linewidth}{!}{%

\begin{tabular}{clcccccc}
\hline
Domaom                       & \multicolumn{1}{c}{Method} & \multicolumn{6}{c}{CKA  For Layers}      \\ \hline
/                            & \multicolumn{1}{c}{/}      & Conv1  & Conv2  & Conv3  & Fc1    & Fc2    & Fc3    \\ \hline
\multirow{5}{*}{MNIST}       & Retrain                    & 0.9997 & 0.9976 & 0.9923 & 0.9741 & 0.9303 & 0.9279 \\
                             & BL1 Repaid Retrain         & 0.9950 & 0.9485 & 0.9712 & 0.4546 & 0.4992 & 0.8646 \\
                             & BL2 FedEraser              & 0.9979 & 0.9909 & 0.9798 & 0.9455 & 0.8830 & 0.8776 \\
                             & BL3 Increase Loss          & 0.9825 & 0.9968 & 0.9876 & 0.9517 & 0.8713 & 0.8925 \\
                             & BL4 Class Pruning          & 0.7892 & 0.9965 & 0.9925 & 0.9756 & 0.9419 & 0.9452 \\ \hline
\multirow{5}{*}{SVHN}        & Retrain                    & 0.9884 & 0.9112 & 0.9114 & 0.8955 & 0.8681 & 0.8519 \\
                             & BL1 Repaid Retrain         & 0.9535 & 0.8856 & 0.4199 & 0.2137 & 0.2757 & 0.7626 \\
                             & BL2 FedEraser              & 0.9801 & 0.8743 & 0.8237 & 0.8026 & 0.8257 & 0.8196 \\
                             & BL3 Increase Loss          & 0.9805 & 0.8788 & 0.8714 & 0.9229 & 0.8856 & 0.8835 \\
                             & BL4 Class Pruning          & 0.9029 & 0.9336 & 0.9509 & 0.9399 & 0.9071 & 0.8936 \\ \hline
\multirow{5}{*}{USPS}        & Retrain                    & 0.9990 & 0.9983 & 0.9927 & 0.9816 & 0.9553 & 0.9279 \\
                             & BL1 Repaid Retrain         & 0.9818 & 0.9935 & 0.9419 & 0.5448 & 0.5537 & 0.8877 \\
                             & BL2 FedEraser              & 0.9966 & 0.9979 & 0.9908 & 0.9775 & 0.9490 & 0.9229 \\
                             & BL3 Increase Loss          & 0.9441 & 0.9944 & 0.9846 & 0.9511 & 0.8816 & 0.8500 \\
                             & BL4 Class Pruning          & 0.8974 & 0.9940 & 0.9912 & 0.9808 & 0.9620 & 0.9363 \\ \hline
\multirow{5}{*}{SynthDigits} & Retrain                    & 0.9975 & 0.9911 & 0.9835 & 0.9466 & 0.8961 & 0.8848 \\
                             & BL1 Repaid Retrain         & 0.9585 & 0.9467 & 0.4720 & 0.2815 & 0.4093 & 0.8287 \\
                             & BL2 FedEraser              & 0.9841 & 0.9894 & 0.9659 & 0.9171 & 0.8721 & 0.8604 \\
                             & BL3 Increase Loss          & 0.9338 & 0.9490 & 0.9766 & 0.9524 & 0.8856 & 0.8952 \\
                             & BL4 Class Pruning          & 0.8314 & 0.9911 & 0.9861 & 0.9612 & 0.9262 & 0.9224 \\ \hline
\multirow{5}{*}{MNIST-M}     & Retrain                    & 0.9938 & 0.9513 & 0.8495 & 0.8236 & 0.7807 & 0.7609 \\ 
                             & BL1 Repaid Retrain         & 0.9787 & 0.9416 & 0.8295 & 0.1376 & 0.3650 & 0.6942 \\
                             & BL2 FedEraser              & 0.9938 & 0.9499 & 0.8525 & 0.8236 & 0.7846 & 0.7628 \\
                             & BL3 Increase Loss          & 0.9719 & 0.9633 & 0.8892 & 0.8639 & 0.7703 & 0.7695 \\
                             & BL4 Class Pruning          & 0.8813 & 0.9420 & 0.8514 & 0.8592 & 0.8490 & 0.8325 \\ \hline
\end{tabular}
}
\end{table}
\begin{table}[!h]

\caption{
CKA for five blocks and three fully connected layer of federated domain unlearning across various methods on DomainNet dataset.}

\label{cka_domainnet}
\renewcommand{\arraystretch}{1.2}
\resizebox{\linewidth}{!}{%

\begin{tabular}{clcccccccc}
\hline
Domaom                     & \multicolumn{1}{c}{Method} & \multicolumn{8}{c}{CKA For Layers}                                    \\  \hline
/                          & \multicolumn{1}{c}{/}      & Block1 & Block2 & Block3 & Block4 & Block5 & Fc1    & Fc2    & Fc3    \\ \hline
\multirow{5}{*}{Clipart}   & Retrain                    & 0.9842 & 0.9789 & 0.9696 & 0.9595 & 0.9140 & 0.8841 & 0.8529 & 0.8494 \\
                           & BL1 Repaid Retrain         & 0.9428 & 0.8966 & 0.7977 & 0.6755 & 0.6672 & 0.5079 & 0.4654 & 0.5167 \\
                           & BL2 FedEraser              & 0.9790 & 0.9697 & 0.9520 & 0.9256 & 0.8717 & 0.8617 & 0.8197 & 0.8110 \\
                           & BL3 Increase Loss          & 0.9831 & 0.9632 & 0.9700 & 0.9624 & 0.9019 & 0.8924 & 0.8562 & 0.8493 \\
                           & BL4 Class Pruning          & 0.9710 & 0.9676 & 0.9626 & 0.9547 & 0.8879 & 0.8931 & 0.8721 & 0.8734 \\ \hline
\multirow{5}{*}{Infograph} & Retrain                    & 0.9616 & 0.9503 & 0.9395 & 0.9171 & 0.8165 & 0.7217 & 0.6087 & 0.6061 \\
                           & BL1 Repaid Retrain         & 0.9149 & 0.8066 & 0.7275 & 0.6678 & 0.6202 & 0.5056 & 0.3117 & 0.3348 \\
                           & BL2 FedEraser              & 0.9678 & 0.9457 & 0.9290 & 0.9123 & 0.8013 & 0.7131 & 0.6041 & 0.5999 \\
                           & BL3 Increase Loss          & 0.9967 & 0.9933 & 0.9871 & 0.9792 & 0.9596 & 0.9463 & 0.9389 & 0.9413 \\
                           & BL4 Class Pruning          & 0.9518 & 0.9437 & 0.9305 & 0.9068 & 0.8508 & 0.8074 & 0.7101 & 0.7538 \\ \hline
\multirow{5}{*}{Painting}  & Retrain                    & 0.9769 & 0.9646 & 0.9585 & 0.9428 & 0.8858 & 0.8822 & 0.8603 & 0.8549 \\
                           & BL1 Repaid Retrain         & 0.8852 & 0.8436 & 0.7791 & 0.7052 & 0.6135 & 0.4911 & 0.5006 & 0.5420 \\
                           & BL2 FedEraser              & 0.9753 & 0.9638 & 0.9361 & 0.9180 & 0.8586 & 0.8618 & 0.8404 & 0.8246 \\
                           & BL3 Increase Loss          & 0.9969 & 0.9911 & 0.9836 & 0.9800 & 0.9764 & 0.9761 & 0.9685 & 0.9673 \\ 
                           & BL4 Class Pruning          & 0.9675 & 0.9680 & 0.9583 & 0.9477 & 0.9320 & 0.9258 & 0.9080 & 0.8772 \\ \hline
\multirow{5}{*}{Quickdraw} & Retrain                    & 0.9990 & 0.9962 & 0.9938 & 0.9915 & 0.9461 & 0.8984 & 0.8558 & 0.8359 \\
                           & BL1 Repaid Retrain         & 0.9824 & 0.9214 & 0.9352 & 0.8780 & 0.8645 & 0.6796 & 0.5316 & 0.4811 \\
                           & BL2 FedEraser              & 0.9982 & 0.9856 & 0.9849 & 0.9805 & 0.9133 & 0.8791 & 0.8372 & 0.8070 \\
                           & BL3 Increase Loss          & 0.9787 & 0.9937 & 0.9796 & 0.9631 & 0.9350 & 0.8927 & 0.8007 & 0.7756 \\
                           & BL4 Class Pruning          & 0.9967 & 0.9980 & 0.9940 & 0.9892 & 0.9564 & 0.9218 & 0.8770 & 0.8647 \\ \hline
\multirow{5}{*}{Real}      & Retrain                    & 0.9813 & 0.9752 & 0.9655 & 0.9606 & 0.9103 & 0.9073 & 0.8977 & 0.8996 \\
                           & BL1 Repaid Retrain         & 0.8896 & 0.8782 & 0.8587 & 0.7635 & 0.6296 & 0.5296 & 0.5163 & 0.5763 \\
                           & BL2 FedEraser              & 0.9694 & 0.9701 & 0.9559 & 0.9416 & 0.8824 & 0.8841 & 0.8781 & 0.8762 \\
                           & BL3 Increase Loss          & 0.9321 & 0.8918 & 0.9396 & 0.9559 & 0.9318 & 0.9100 & 0.8506 & 0.8529 \\
                           & BL4 Class Pruning          & 0.9540 & 0.9596 & 0.9595 & 0.9459 & 0.8866 & 0.8931 & 0.8840 & 0.8882 \\ \hline
\multirow{5}{*}{Sketch}    & Retrain                    & 0.9900 & 0.9807 & 0.9655 & 0.9555 & 0.8764 & 0.8351 & 0.8199 & 0.8250 \\
                           & BL1 Repaid Retrain         & 0.9267 & 0.8606 & 0.7982 & 0.7044 & 0.6064 & 0.4493 & 0.3677 & 0.4231 \\
                           & BL2 FedEraser              & 0.9716 & 0.9666 & 0.9471 & 0.9258 & 0.8351 & 0.7896 & 0.7795 & 0.7799 \\
                           & BL3 Increase Loss          & 0.9982 & 0.9928 & 0.9872 & 0.9884 & 0.9770 & 0.9667 & 0.9629 & 0.9642 \\
                           & BL4 Class Pruning          & 0.9729 & 0.9839 & 0.9818 & 0.9723 & 0.9196 & 0.9082 & 0.9029 & 0.9083 \\ \hline
\end{tabular}
}
\end{table}

\begin{figure}[!b]
\centering
\includegraphics[width=1\linewidth]{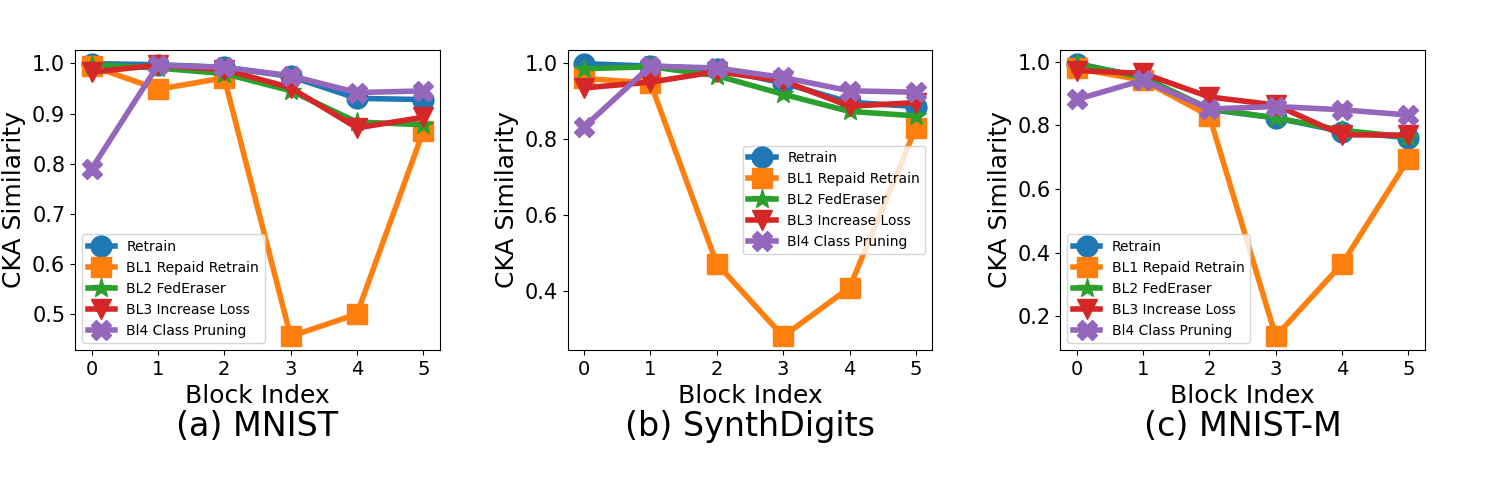}
\vspace{-25pt}
\caption{ CKA analysis of layer representations before and after unlearning the target domain in Domain-Digital. We visualize three domains: (a) MNIST, (b) SynthDigits, and (c) MNIST-M.}
\label{fig_cka_dig}
\end{figure}

\begin{figure}[!b]
\centering
\includegraphics[width=1\linewidth]{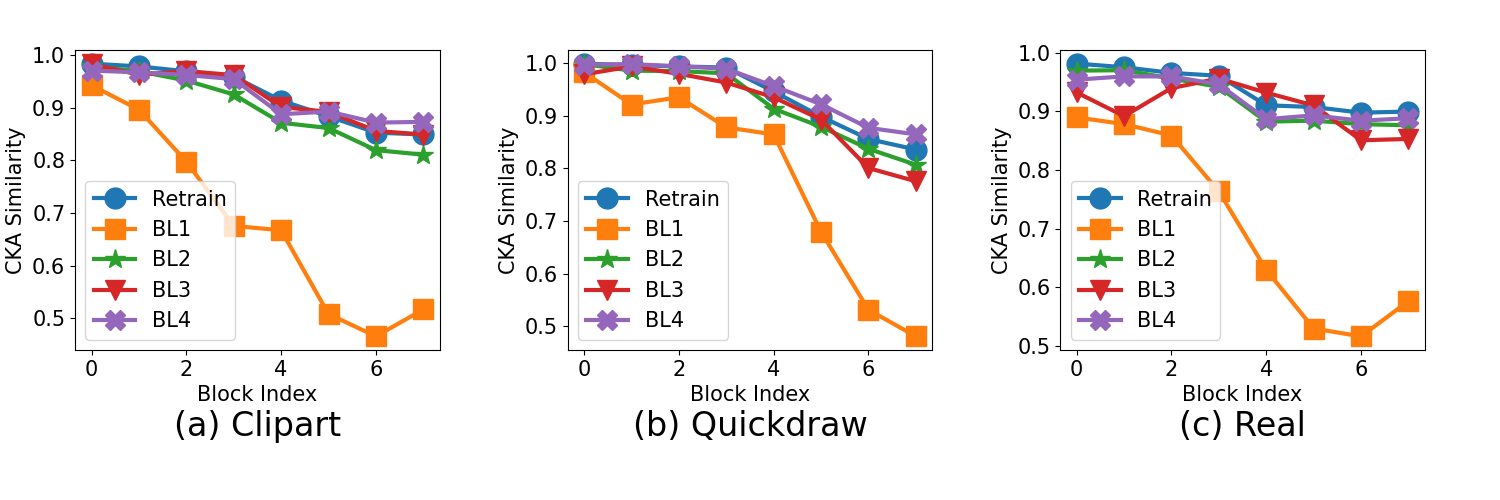}
\vspace{-25pt}
\caption{CKA analysis of layer representations before and after unlearning the target domain in DomainNet. We visualize the rest three domains: (a) Clipart, (b) Quickdraw, and (c) Real.}
\label{fig_cka_dom}
\end{figure}

\subsection{Federated Domain Unlearning and Hidden Layer Representations}
We employ the Centered Kernel Alignment (CKA) metric \cite{kornblith2019similarity}, a tool for assessing the similarity between neural network representations. CKA quantifies the similarity between two neural networks by computing the inner product between their centered kernel matrices. This approach provides a measure of how much common information is retained between the networks, thereby shedding light on the extent of information preservation or loss during unlearning.
In our experimental setup, we utilize linear CKA to analyze the similarity of the output features produced by two models before and after the unlearning process. Given a dataset \(D_{cka}\), we extract feature matrices \(Z_1\) and \(Z_2\) from the two models, respectively. The linear CKA similarity between two representations \(X\) and \(Y\) is calculated using the following equation:
\[
CKA(X,Y) = \frac{||X^TY||^2_F}{||X^TX||^2_F \cdot ||Y^TY||^2_F},
\]
where \(|| \cdot ||_F\) denotes the Frobenius norm. This formula yields a similarity score ranging from 0 (indicating no similarity) to 1 (indicating identical representations), thereby enabling a quantitative assessment of how similar the output features of the same layer are across two models.

All the results of Centered Kernel Alignment (CKA) across multiple target
domains from Domain-Digital and DomainNet dataset, comparing various unlearning methods with the comprehensive learning model, were displayed in Tables \ref{cka_digit} and Tables \ref{cka_domainnet}. Furthermore, we visualized the remaining three domains of DomainNet in Figure \ref{fig_cka_dig} and parts of the Domain-Digital in Figure \ref{fig_cka_dom}. The results on Domain-Digital are found to be similar to those on DomainNet. However, a notable difference is that Class-Discriminative Pruning has a significant impact on the first convolutional kernel of the network used for training Domain-Digital, which has three convolutional layers. We also analyzed the CKA scores of all convolutional layers of VGG16 and found similar results of Class-Discriminative Pruning.

\begin{table}[!h]

\caption{
Subspace similarity for three convs of federated domain unlearning across various methods on Domain-Digital dataset.}

\label{subspace_similarity_digit}
\renewcommand{\arraystretch}{1.2}
\resizebox{\linewidth}{!}{%

\begin{tabular}{clccc}
\hline
Domaom                       & \multicolumn{1}{c}{Method} & \multicolumn{3}{c}{Subspace  Similarity For Layers} \\  \hline
/                            & \multicolumn{1}{c}{/}      & Conv1           & Conv2           & Conv3           \\ \hline
\multirow{5}{*}{MNIST}       & Retrain                    & 0.9405          & 0.4643          & 0.4721          \\
                             & BL1 Repaid Retrain         & 0.8567          & 0.2033          & 0.3834          \\
                             & BL2 FedEraser              & 0.8978          & 0.5045          & 0.4895          \\
                             & BL3 Increase Loss          & 0.6421          & 0.7798          & 0.7387          \\ 
                             & BL4 Class Pruning          & 0.0252          & 0.4901          & 0.4629          \\ \hline
\multirow{5}{*}{SVHN}        & Retrain                    & 0.7170          & 0.3751          & 0.3967          \\
                             & BL1 Repaid Retrain         & 0.5605          & 0.2628          & 0.0245          \\
                             & BL2 FedEraser              & 0.6917          & 0.3379          & 0.3783          \\
                             & BL3 Increase Loss          & 0.5703          & 0.6731          & 0.6503          \\
                             & BL4 Class Pruning          & 0.0117          & 0.4940          & 0.3704          \\ \hline
\multirow{5}{*}{USPS}        & Retrain                    & 0.9054          & 0.4820          & 0.4371          \\
                             & BL1 Repaid Retrain         & 0.7042          & 0.3971          & 0.2987          \\
                             & BL2 FedEraser              & 0.8731          & 0.4976          & 0.4666          \\
                             & BL3 Increase Loss          & 0.2641          & 0.7718          & 0.6912          \\
                             & BL4 Class Pruning          & 0.0178          & 0.5805          & 0.4695          \\ \hline
\multirow{5}{*}{SynthDigits} & Retrain                    & 0.8688          & 0.4748          & 0.5183          \\
                             & BL1 Repaid Retrain         & 0.5500          & 0.3058          & 0.0011          \\
                             & BL2 FedEraser              & 0.8175          & 0.4952          & 0.5193          \\
                             & BL3 Increase Loss          & 0.3175          & 0.7013          & 0.7874          \\
                             & BL4 Class Pruning          & 0.0058          & 0.5588          & 0.4355          \\ \hline
\multirow{5}{*}{MNIST-M}     & Retrain                    & 0.7041          & 0.5205          & 0.4664          \\
                             & BL1 Repaid Retrain         & 0.7563          & 0.4249          & 0.3608          \\
                             & BL2 FedEraser              & 0.7030          & 0.5121          & 0.4739          \\
                             & BL3 Increase Loss          & 0.5553          & 0.8213          & 0.6629          \\
                             & BL4 Class Pruning          & 0.0148          & 0.5250          & 0.4045          \\ \hline 
\end{tabular}
}
\end{table}
\begin{table}[]

\caption{
Subspace similarity for five blocks of federated domain unlearning across various methods on DomainNet dataset.}

\label{subspace_similarity_domainnet}
\renewcommand{\arraystretch}{1.2}
\resizebox{\linewidth}{!}{%

\begin{tabular}{clccccc}
\hline
Domaom                     & \multicolumn{1}{c}{Method} & \multicolumn{5}{c}{Subspace  Similarity For Layers} \\ \hline
/                          & \multicolumn{1}{c}{/}      & Block1   & Block2   & Block3   & Block4   & Block5  \\ \hline
\multirow{5}{*}{Clipart}   & Retrain                    & 0.4368   & 0.3365   & 0.2873   & 0.2611   & 0.2219  \\
                           & BL1 Repaid Retrain         & 0.3077   & 0.3595   & 0.2440   & 0.1829   & 0.0828  \\
                           & BL2 FedEraser              & 0.3487   & 0.3793   & 0.3008   & 0.2793   & 0.2478  \\
                           & BL3 Increase Loss          & 0.6765   & 0.3507   & 0.4163   & 0.5783   & 0.5189  \\
                           & BL4 Class Pruning          & 0.1645   & 0.2276   & 0.2678   & 0.2516   & 0.1001  \\ \hline
\multirow{5}{*}{Infograph} & Retrain                    & 0.4121   & 0.3048   & 0.2017   & 0.1377   & 0.1460  \\
                           & BL1 Repaid Retrain         & 0.2657   & 0.2634   & 0.1841   & 0.1053   & 0.0919  \\
                           & BL2 FedEraser              & 0.3873   & 0.2077   & 0.2001   & 0.1500   & 0.1458  \\
                           & BL3 Increase Loss          & 0.9867   & 0.9669   & 0.9337   & 0.9071   & 0.8728  \\
                           & BL4 Class Pruning          & 0.2464   & 0.2497   & 0.2563   & 0.1744   & 0.0582  \\ \hline
\multirow{5}{*}{Painting}  & Retrain                    & 0.4869   & 0.3445   & 0.2108   & 0.1940   & 0.1570  \\
                           & BL1 Repaid Retrain         & 0.2584   & 0.2474   & 0.1410   & 0.0932   & 0.0714  \\
                           & BL2 FedEraser              & 0.4059   & 0.3096   & 0.1981   & 0.1821   & 0.1965  \\
                           & BL3 Increase Loss          & 0.9759   & 0.9420   & 0.7803   & 0.8632   & 0.9225  \\
                           & BL4 Class Pruning          & 0.1607   & 0.2228   & 0.2196   & 0.2008   & 0.0854  \\ \hline
\multirow{5}{*}{Quickdraw} & Retrain                    & 0.6189   & 0.4184   & 0.2596   & 0.1829   & 0.1274  \\
                           & BL1 Repaid Retrain         & 0.4428   & 0.3551   & 0.2701   & 0.1617   & 0.0571  \\
                           & BL2 FedEraser              & 0.5350   & 0.4446   & 0.2932   & 0.2063   & 0.1347  \\
                           & BL3 Increase Loss          & 0.6195   & 0.4756   & 0.2230   & 0.3650   & 0.4007  \\
                           & BL4 Class Pruning          & 0.3796   & 0.4469   & 0.3286   & 0.2962   & 0.1708  \\ \hline
\multirow{5}{*}{Real}      & Retrain                    & 0.3708   & 0.2344   & 0.1648   & 0.1785   & 0.1754  \\
                           & BL1 Repaid Retrain         & 0.2309   & 0.1907   & 0.1091   & 0.0746   & 0.0577  \\
                           & BL2 FedEraser              & 0.3668   & 0.2767   & 0.1935   & 0.1903   & 0.1727  \\
                           & BL3 Increase Loss          & 0.2993   & 0.1553   & 0.3644   & 0.6277   & 0.5379  \\
                           & BL4 Class Pruning          & 0.1963   & 0.1835   & 0.2130   & 0.1962   & 0.1090  \\ \hline
\multirow{5}{*}{Sketch}    & Retrain                    & 0.3961   & 0.3905   & 0.2437   & 0.1951   & 0.1923  \\
                           & BL1 Repaid Retrain         & 0.2610   & 0.3528   & 0.2243   & 0.1444   & 0.0866  \\
                           & BL2 FedEraser              & 0.3416   & 0.3858   & 0.2560   & 0.2203   & 0.2136  \\
                           & BL3 Increase Loss          & 0.9672   & 0.9549   & 0.8586   & 0.8229   & 0.9066  \\
                           & BL4 Class Pruning          & 0.3364   & 0.3515   & 0.2581   & 0.2840   & 0.1334  \\ \hline
\end{tabular}
}
\end{table}

\begin{figure}[!t]
\centering
\includegraphics[width=1\linewidth]{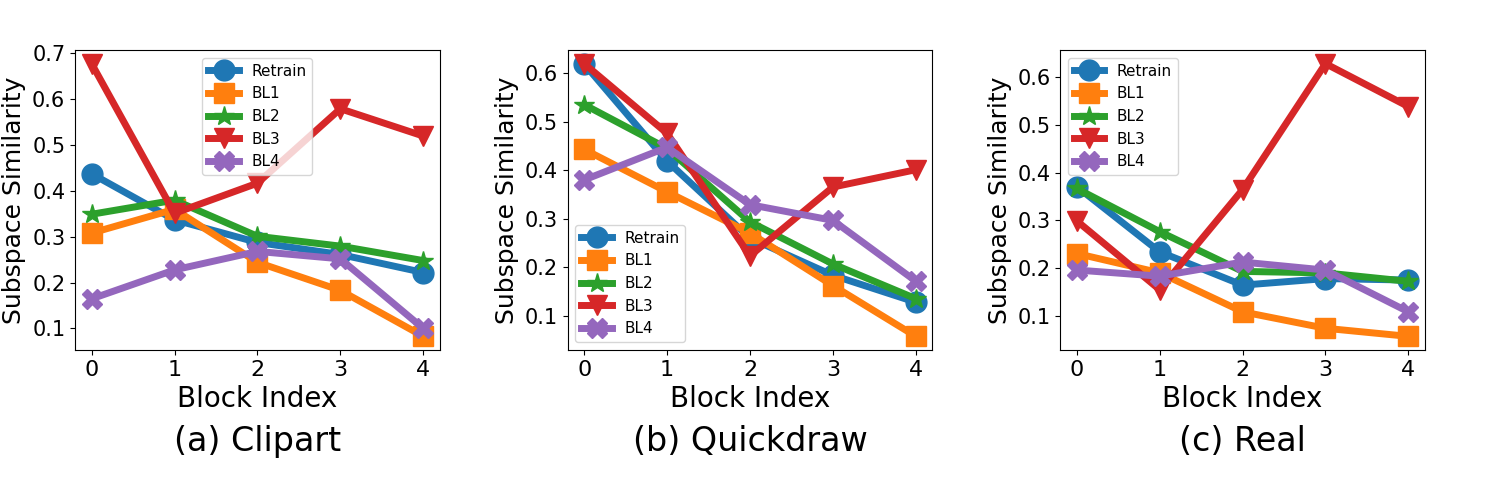}
\vspace{-25pt}
\caption{ Comparative analysis of subspace similarity in feature extractors before and after unlearning in the target domain of DomainNet. We visualize the rest three domains: (a) Clipart, (b) Quickdraw, and (c) Real.}
\label{fig_sub_dom}
\end{figure}

\begin{figure}[!t]
\centering
\includegraphics[width=1\linewidth]{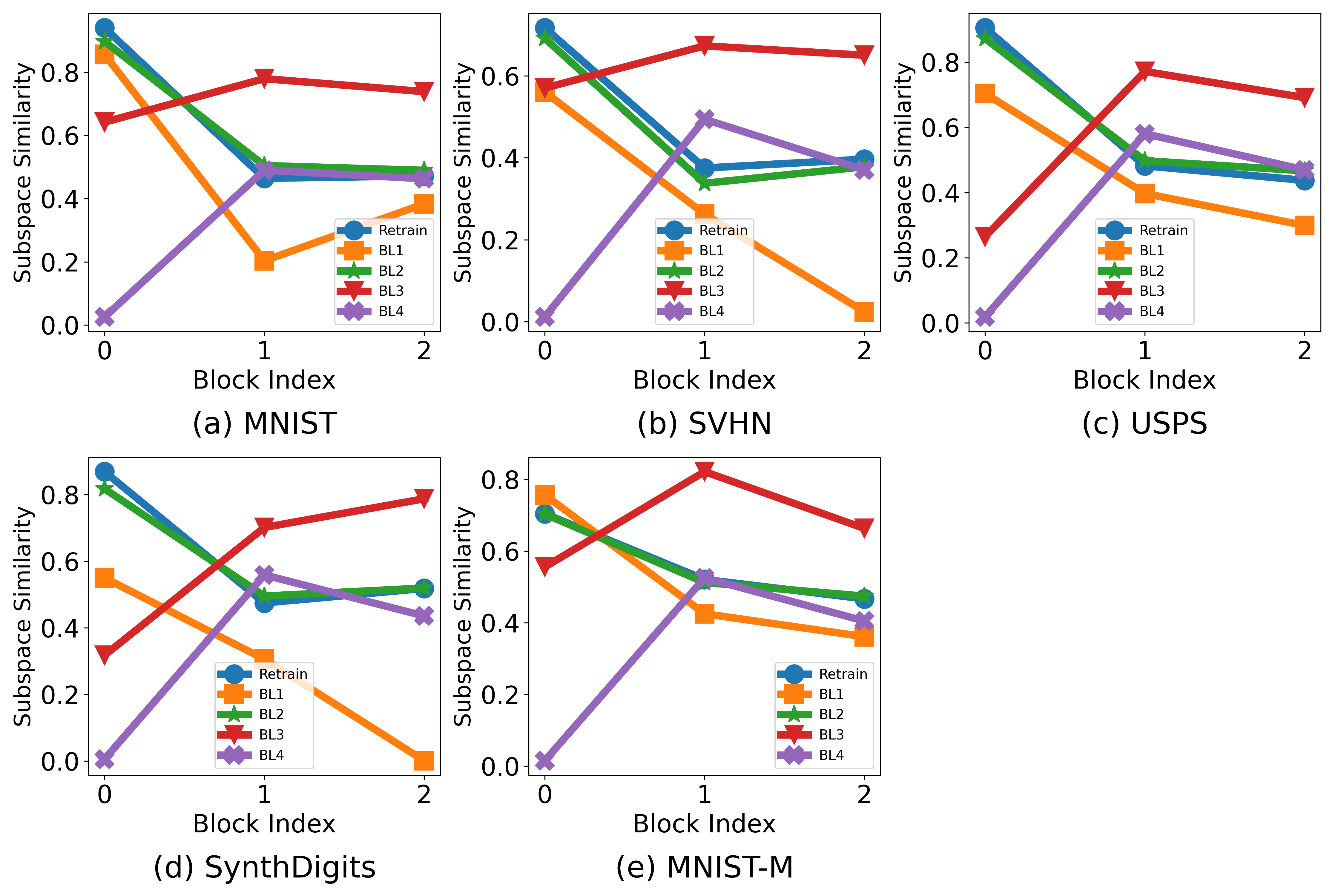}
\vspace{-25pt}
\caption{ Comparative analysis of subspace similarity in feature extractors before and after unlearning in the target domain of Domain-Digital.}
\label{fig_sub_digit}
\end{figure}

\subsection{Feature Reuse}
To further investigate how the representations of lower and higher layers evolve during unlearning, we conduct the subspace similarity analysis on the unlearned models with different unlearning methods. Let \( A \in \mathbb{R}^{n \times m} \) represent the centered layer activation matrix with \( n \) examples and \( m \) neurons. We determine the PCA decomposition of \( A \), which involves computing the eigenvectors \( (e_1, e_2, ...) \) and the corresponding eigenvalues \( (\delta_1, \delta_2, ...) \) of the matrix \( A^TA \). Let \( E_k \) denote the matrix composed of the first \( k \) principal components, with \( e_1, ..., e_k \) as its columns, and let \( G_k \) be the analogous matrix derived from another activation matrix \( B \). We then compute the subspace similarity for the top \( k \) components as:
\vspace{-6pt}
\begin{equation}
    \text{SubspaceSim}_k(A, B) = ||G_k^T \cdot E_k||_F^2 
    \vspace{-4pt}
\end{equation}
This metric quantifies the congruence of the subspaces spanned by \( (e_1, ..., e_k) \) and \( (g_1, ..., g_k) \). For instance, if \( A \) and \( B \) are the layer activation matrices corresponding to different tasks, then \( \text{SubspaceSim}_k \) evaluates the similarity in how the network encodes the top \( k \) features for those tasks. 

All the results of the subspace similarity of feature extractors before and after the application of various unlearning methods in the Domain-Digital and DomainNet were displayed in Tables \ref{subspace_similarity_digit} and Tables \ref{subspace_similarity_domainnet}. Furthermore, we visualized the remaining three domains of DomainNet in Figure \ref{fig_sub_dom} and all domains of the Domain-Digital in Figure \ref{fig_sub_digit}.

\begin{figure}[!t]
\centering
\includegraphics[width=1\linewidth]{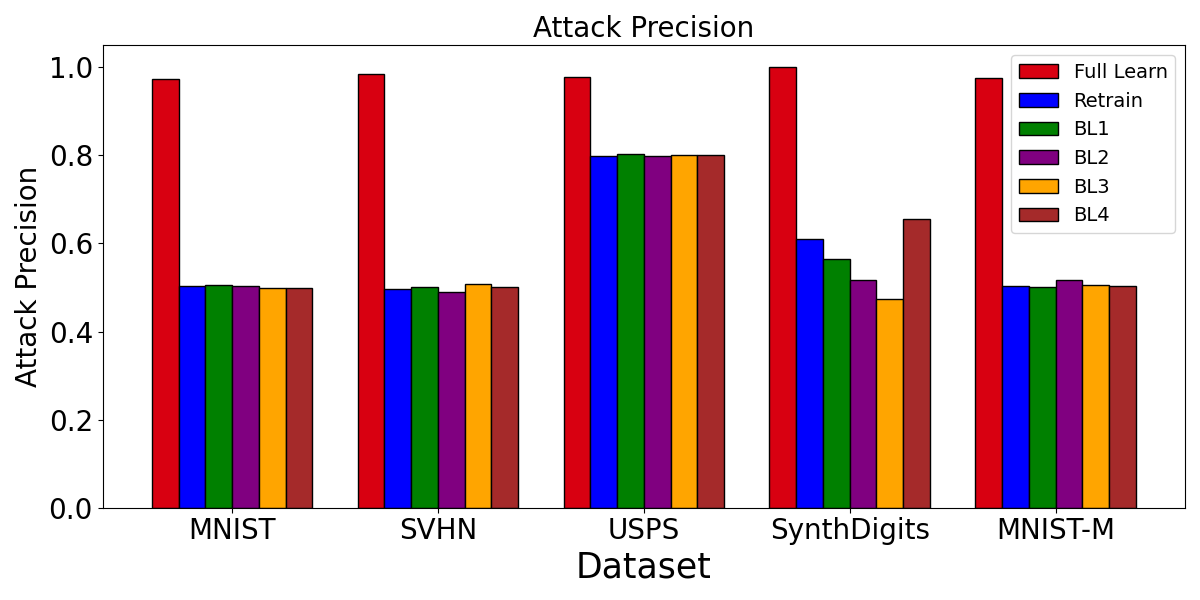}
\vspace{-25pt}
\caption{Attack precision of membership inference attacks.}
\label{fig_mia}
\end{figure}

\begin{figure}[!t]
\centering
\includegraphics[width=1\linewidth]{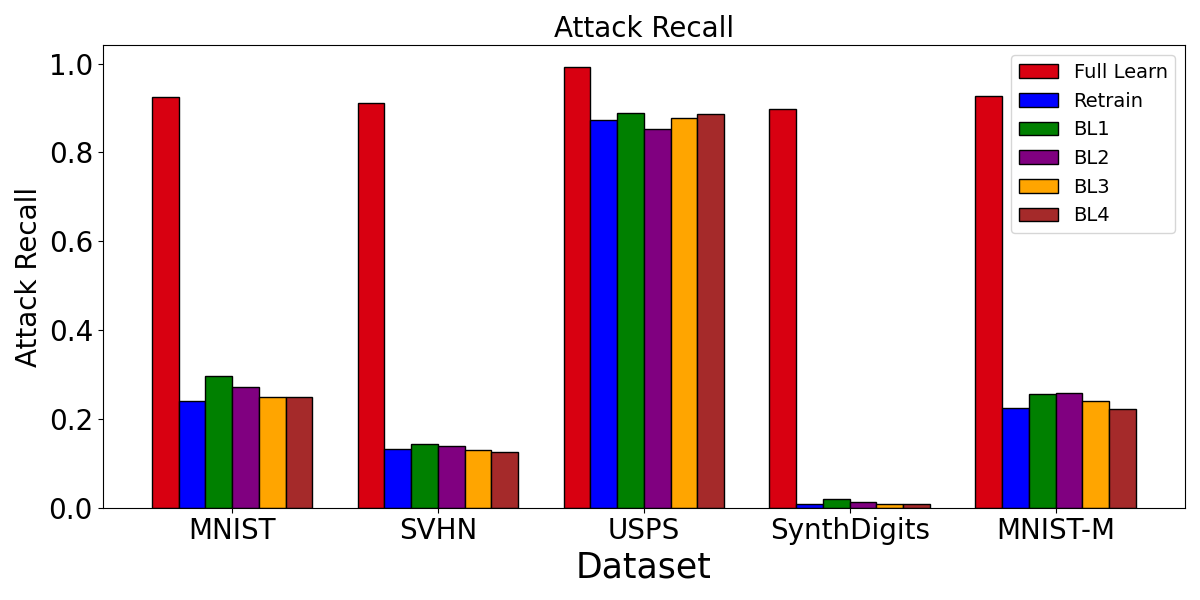}
\vspace{-25pt}
\caption{Attack recall of membership inference attacks.}
\label{fig_mia_recall}
\end{figure}

\subsection{Membership Inference Attack}
We perform Membership Inference Attack \cite{shokri2017membership} (MIA) experiments, employing the strategy of shadow model training to extract data for the purpose of constructing an attack classifier. Utilizing fully trained models that encompass all domains as shadow models, we conduct attacks on models from which certain domains have been unlearned through various unlearning methods. We measure both the attack accuracy and attack recall which demonstrate the amount of information about the data in a domain that remains in the unlearned model. The ideal unlearning method would minimize both accuracy and recall, indicating the attack model's difficulty in distinguishing whether the unlearned domain had participated in federated learning. From Figures \ref{fig_mia} and Figures \ref{fig_mia_recall}, it can be observed that there are significant differences in sensitivity and specificity across different domains. USPS exhibits high accuracy and recall in attacks, whereas SVHN and SynthDigitls show lower values, especially in attack recall, with SynthDigitls being notably low. Additionally, across most domains, various unlearning methods slightly higher than retrain, both in attack precision and recall.

\begin{table}[]

\caption{Evaluation of our verification method for domain unlearning and backdoor attack across various methods in Office-Caltech10 dataset.}
\label{bkd_office_train}
\vspace{-10pt}
\resizebox{\linewidth}{!}{

\begin{tabular}{ccccccc}
\hline
\multirow{2}{*}{Doamin}  & \multirow{2}{*}{Method} & \multicolumn{5}{c}{\multirow{2}{*}{\textbf{Train Accuracy  For BaseLine}}} \\
                         &                         & \multicolumn{5}{c}{}                                                  \\ \hline
/                        & /                       & Retrain       & BL1         & BL2         & BL3         & BL4         \\ \hline
\multirow{3}{*}{Amazon}  & Clean                   & 61.98         & 43.19       & 51.96       & 71.04       & 67.23       \\
                         & Backdoor                & 9.04          & 0.87        & 4.08        & 8.54        & 2.92        \\
                         & Ours                    & 0             & 0           & 0           & 2.48        & 7.83        \\ \hline
\multirow{3}{*}{Caltech} & Clean                   & 33.32         & 30.94       & 37.31       & 91.85       & 43.16       \\
                         & Backdoor                & 4.64          & 4.64        & 2.32        & 8.42        & 14.69       \\
                         & Ours                    & 0             & 0           & 0           & 18.93       & 19.04       \\ \hline
\multirow{3}{*}{Dslr}    & Clean                   & 77.28         & 70.88       & 68.2        & 80.64       & 79.36       \\
                         & Backdoor                & 5.17          & 3.45        & 8.62        & 6.28        & 5.17        \\
                         & Ours                    & 0             & 0           & 0           & 28.00          & 13.60        \\ \hline
\multirow{3}{*}{Webcam}  & Clean                   & 63.31         & 61.44       & 56.78       & 65.25       & 63.14       \\
                         & Backdoor                & 5.66          & 1.89        & 4.75        & 7.13        & 22.64       \\
                         & Ours                    & 0             & 12.71       & 8.89       & 27.96       & 24.15       \\ \hline
\end{tabular}
}
\vspace{-10pt}
\end{table}
\begin{table}[]

\caption{Evaluation results of our verification method and backdoor attacks on original model performance in Office-Caltech10 dataset.}
\vspace{-10pt}
\label{bkd_office_test}
\resizebox{\linewidth}{!}{

\begin{tabular}{cccccc}
\hline
Domain                   & Method   & \multicolumn{4}{c}{\textbf{Test  Accuracy  For All  Domain}} \\ \hline
/                        & /        & Amazon       & Caltech       & Dslr       & Webcam      \\ \hline
/                        & Clean    & 75.52        & 74.67         & 84.38      & 86.44       \\ \hline
\multirow{2}{*}{Amazon}  & Backdoor & 6.25         & 1.78          & -3.12      & -10.17      \\
                         & Ours     & 5.73         & 2.23          & -3.12      & -10.17      \\ \hline
\multirow{2}{*}{Caltech} & Backdoor & 23.96        & 17.78         & 6.26       & 5.08        \\
                         & Ours     & 3.64         & 5.78          & 6.26       & 8.47        \\ \hline
\multirow{2}{*}{Dslr}    & Backdoor & 7.17         & 4.67          & 3.63       & 6.30         \\ 
                         & Ours     & 6.77         & 2.67          & 0.38       & 2.39        \\ \hline
\multirow{2}{*}{Webcam}  & Backdoor & 3.92         & 0.45          & -0.12      & 3.39        \\
                         & Ours     & 2.60          & 1.34          & -2.87      & -6.78       \\ \hline
\end{tabular}
}
\vspace{-10pt}
\end{table}
\begin{table}[!h]

\caption{Evaluation results of our verification method and backdoor attacks on original model performance in DomainNet dataset.}
\vspace{-10pt}
\label{bkd_DomainNet_test}
\resizebox{\linewidth}{!}{

\begin{tabular}{cccccccc}
\hline
Domain                     & Method   & \multicolumn{6}{c}{\textbf{Test  Accuracy  For All  Domain}}      \\ \hline
/                          & /        & Clipart & Infograph & Infograph & Quickdraw & Real  & Sketch \\ \hline
/                          & Clean    & 73.95   & 38.05     & 61.39     & 61.30      & 67.79 & 66.06  \\ \hline
\multirow{2}{*}{Clipart}   & Backdoor & 7.22    & 2.13      & 1.94      & -1.90      & 0.25  & 2.16   \\
                           & Ours     & 4.18    & 1.06      & -1.61     & 0.60       & -2.14 & -4.70   \\ \hline
\multirow{2}{*}{Infograph} & Backdoor & -0.57   & 3.50       & 1.29      & -5.50      & -0.16 & 1.62   \\
                           & Ours     & -2.29   & -0.15     & -5.09     & -2.50      & -5.09 & -5.42  \\ \hline
\multirow{2}{*}{Painting}  & Backdoor & -0.19   & 0.76      & 1.78      & 1.50       & -1.48 & 2.16   \\
                           & Ours     & -2.10    & 1.67      & 0.16      & -2.40      & -2.38 & -3.25  \\ \hline
\multirow{2}{*}{Quickdraw} & Backdoor & 1.52    & 1.37      & -0.32     & -0.60      & -1.89 & 2.88   \\ 
                           & Ours     & -0.19   & -1.98     & -2.42     & 4.50       & -2.14 & -2.17  \\ \hline
\multirow{2}{*}{Real}      & Backdoor & -0.95   & 0.15      & 1.13      & -4.10      & 0.58  & 0.54   \\
                           & Ours     & 3.23    & 1.82      & -4.36     & -2.50      & -0.82 & -4.16  \\ \hline
\multirow{2}{*}{Sketch}    & Backdoor & 3.42    & 1.82      & 1.13      & -2.96     & -0.74 & 4.87   \\
                           & Ours     & 3.23    & 1.06      & -1.13     & -1.40      & -2.22 & 4.51   \\ \hline
\end{tabular}
}
\vspace{-10pt}
\end{table}
\begin{table}[]

\caption{Evaluation of our verification method for domain unlearning and backdoor attack across various methods in DomainNet dataset.}
\label{bkd_DomainNet_train}
\vspace{-10pt}
\resizebox{\linewidth}{!}{

\begin{tabular}{ccccccc}
\hline

\multirow{2}{*}{Doamin}    & \multirow{2}{*}{Method} & \multicolumn{5}{c}{\multirow{2}{*}{\textbf{Train Accuracy  For BaseLine}}} \\
                           &                         & \multicolumn{5}{c}{}                                                  \\ \hline
/                          & /                       & Retrain       & BL1         & BL2         & BL3         & BL4         \\ \hline
\multirow{3}{*}{Clipart}   & Clean                   & 68.19         & 51.88       & 67.38       & 81.93       & 76.18       \\
                           & Backdoor                & 1.84          & 5.06        & 3.68        & 3.54        & 3.91        \\
                           & Ours                    & 0             & 7.63        & 26.50        & 25.80        & 6.40         \\ \hline
\multirow{3}{*}{Infograph} & Clean                   & 36.04         & 28.01       & 33.48       & 70.00          & 37.38       \\
                           & Backdoor                & 7.71          & 9.42        & 11.35       & 5.41        & 10.06       \\
                           & Ours                    & 0             & 0           & 17.90         & 65.60        & 12.80        \\ \hline
\multirow{3}{*}{Painting}  & Clean                   & 65.57         & 49.07       & 65.00          & 85.69       & 70.42       \\
                           & Backdoor                & 3.79          & 10.49       & 5.13        & 4.77        & 6.92        \\
                           & Ours                    & 0             & 33.70        & 9.60         & 54.80        & 11.30        \\ \hline
\multirow{3}{*}{Quickdraw} & Clean                   & 50.45         & 40.51       & 48.60        & 54.73       & 54.78       \\
                           & Backdoor                & 8.33          & 6.36        & 10.31       & 0           & 7.68        \\
                           & Ours                    & 0             & 0           & 30.70        & 39.00          & 36.20        \\ \hline
\multirow{3}{*}{Real}      & Clean                   & 72.94         & 55.16       & 71.37       & 81.93       & 74.47       \\
                           & Backdoor                & 1.35          & 0.90         & 0.90         & 0.45        & 3.60         \\
                           & Ours                    & 0             & 4.30         & 29.40         & 43.60        & 21.70        \\ \hline
\multirow{3}{*}{Sketch}    & Clean                   & 63.15         & 39.37       & 58.16       & 89.92       & 73.89       \\
                           & Backdoor                & 2.68          & 4.25        & 2.91        & 2.68        & 2.24        \\
                           & Ours                    & 0             & 5.60        & 18.70        & 14.00          & 19.30        \\ \hline
\end{tabular}}
\vspace{-10pt}

\end{table}

\begin{figure}[!t]
\centering
\includegraphics[width=1\linewidth]{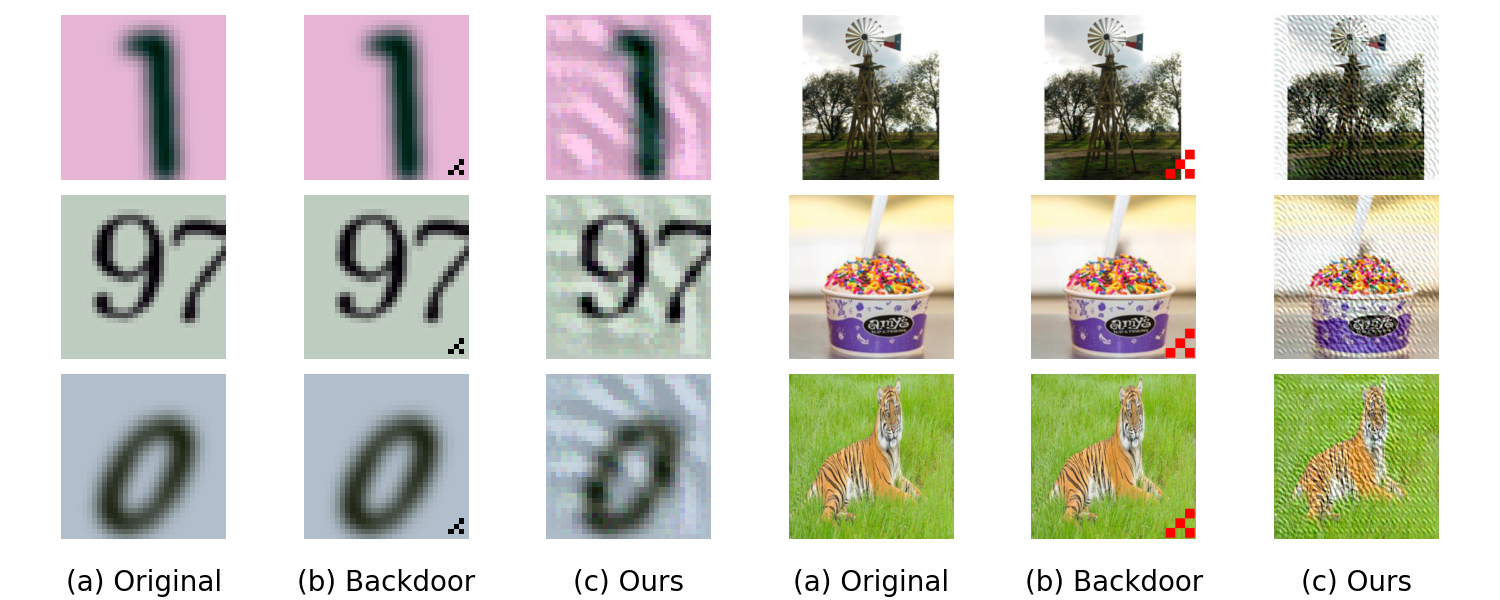}
\caption{The differences between our verification method and backdoor attack. On the left are images from the Domain-Digital dataset, and on the right are images from the DomainNet dataset.}
\label{fig_bkd_vis}
\end{figure}

\begin{figure}[!t]
\centering
\includegraphics[width=1\linewidth]{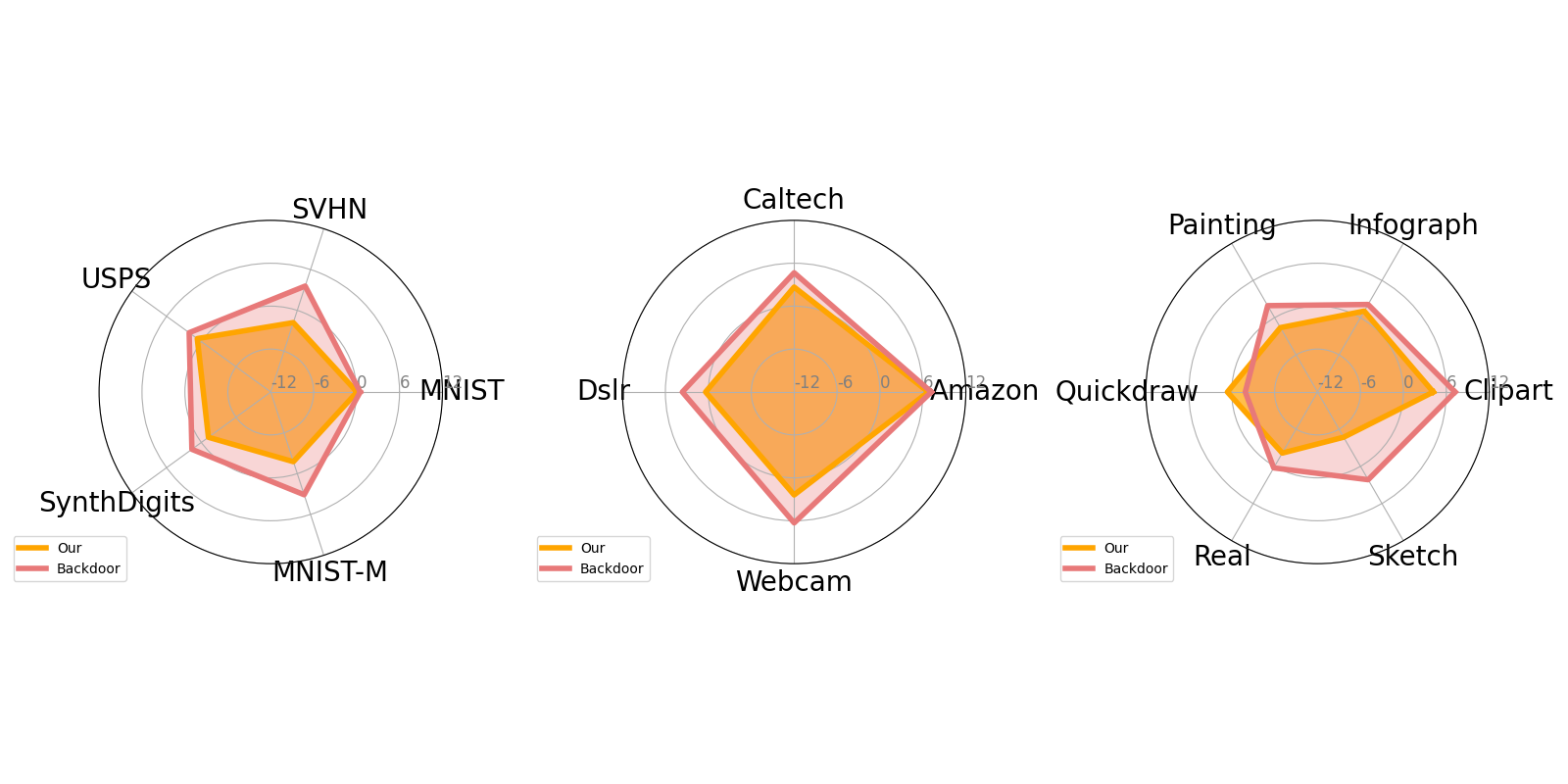}
\vspace{-35pt}
\caption{Test accuracy for backdoor and our verification in three datasets. The smaller the area of the polygon, the lesser the impact on performance loss.}
\label{fig_bkd}
\vspace{-10pt}
\end{figure}

\subsection{Our Validation Results}
We conduct experimental comparisons between traditional backdoor methods, which involve adding pixels or patterns, and our proposed verification method. We can see the images in Figures \ref{fig_bkd_vis}. The detials of the efficacy of our verification method in terms of domain sensitivity and specificity were shown in Tables \ref{bkd_office_train} and Tables \ref{bkd_office_test} for Office-Caltech dataset, Tables \ref{bkd_DomainNet_test} and Tables \ref{bkd_DomainNet_train} for DomainNet dataset. It is evident that compared to backdoor attacks, our verification method demonstrated a smaller performance loss. Figure \ref{fig_bkd} selects one domain from each of the three datasets to more distinctly showcase the superiority of our verification method.


\begin{table}[!h]
\caption{Evaluation results of our verification method on original model performance under different parameters of Office-Caltech10 Caltech  Dataset.}

\label{bkd_setting_test}
\resizebox{\linewidth}{!}{

\begin{tabular}{ccccc}
\hline
Setting       & \multicolumn{4}{c}{Test  Accuracy  For All  Domain} \\ \hline
$\epsilon$ / $\mu$,${\lambda}$       & Amazon     & Caltech    & Dslr     & Webcam    \\ \hline
Clean         & 75.52      & 74.67      & 84.38    & 86.44     \\ \hline
$\epsilon$=0.3/$\mu$,${\lambda}$=0.5 & 71.88      & 68.89      & 78.12    & 77.97     \\ \hline
$\epsilon$=0.1         & 66.67      & 69.78      & 81.25    & 89.83     \\ \hline
$\epsilon$=0.6         & 70.31      & 69.33      & 68.75    & 71.19     \\ \hline
$\mu$,${\lambda}$=0.3,0.7   & 61.46      & 63.56      & 81.25    & 74.75     \\ \hline
$\mu$,${\lambda}$=0.7,0.3   & 72.40       & 72.44      & 81.25    & 84.75     \\ \hline
\end{tabular}
}
\end{table}

\begin{table}[!h]
\caption{ Validation results of our verification method under different parameters of Office-Caltech10 Caltech  Dataset.}

\label{bkd_setting_train}
\resizebox{\linewidth}{!}{

\begin{tabular}{ccccccc}
\hline
Setting       & \multicolumn{6}{c}{Train Accuracy  For BaseLine} \\ \hline
${\epsilon}$ / $\mu$,${\lambda}$       & Full  & Retrain & BL1 & BL2 & BL3   & BL4   \\ \hline
${\epsilon}$=0.3/$\mu$,${\lambda}$=0.5 & 99.89 & 0       & 0   & 0   & 18.93 & 19.04 \\ \hline
${\epsilon}$=0.1         & 98.89 & 0       & 0   & 0   & 37.86 & 8.35  \\ \hline
${\epsilon}$=0.6         & 89.64 & 0       & 0   & 5.97   & 52.11 & 18.37 \\ \hline
$\mu$,${\lambda}$=0.3,0.7   & 99.89 & 0       & 0   & 7.56   & 40.31 & 9.02  \\ \hline
$\mu$,${\lambda}$=0.7,0.3   & 100.00     & 0       & 0   & 0   & 41.65 & 5.79  \\ \hline
\end{tabular}
}
\end{table}
\subsection{Ablation Study}
We examine the impact of hyperparameters \(\epsilon\), \(\mu\), and \(\lambda\) on our verification method. Aside from the Domain-Digit datasets, we also conduct experiments on the Office-Caltech10 dataset, with the results shown in Table \ref{bkd_setting_test} and \ref{bkd_setting_train}. It is evident that excessively high values of \(\epsilon\) and \(\lambda\) can significantly impair performance, while too low values of \(\epsilon\) may lead to weaker verification effects. 
Parameters \(\mu\) and \(\lambda\) balance the loss contributions from clean and marker-injected data; a higher \(\mu\) compared to \(\lambda\) shifts the model's focus towards optimizing performance on clean data, enabling rapid convergence to the performance of an unmodified classifier. This study emphasizes the need for precise calibration of these parameters to ensure balanced learning and effective unlearning verification.

\end{document}